\documentclass[12pt,authoryear]{elsarticle}

\usepackage{amssymb}

\usepackage{times}
\usepackage{latexsym}

\usepackage{graphicx}
\usepackage{epstopdf}
\usepackage{booktabs}
\usepackage{xspace}
\usepackage{multirow}
\usepackage[tight]{subfigure}
\usepackage{enumitem}
\usepackage{hhline}
\usepackage{arydshln}
\usepackage{rotating}
\usepackage{url}
\usepackage{CJKutf8}
\usepackage{textcomp}
\usepackage{amsmath}
\usepackage{microtype}
\usepackage{rotating}

\newcommand{\ie}{i.e.,\xspace}

\newcommand{\furl}[1]{\footnote{\scriptsize \url{#1}}}

\newcommand{\ftext}[1]{\footnote{\scriptsize #1}}

\journal{Knowledge-based Systems}

\begin{document}

\begin{frontmatter}

\title{Marshall-Olkin Power-Law Distributions in Length-Frequency of Entities}

\author[inst1,inst3]{Xiaoshi Zhong}
\ead{xszhong@bit.edu.cn}
\author[inst1]{Xiang Yu}
\ead{yuxiang@bit.edu.cn}

\affiliation[inst1]{organization={School of Computer Science and Technology, Beijing Institute of Technology},
            country={China}}

\affiliation[inst3]{organization={State Key Laboratory of Software Development Environment, Beihang University},
            country={China}}

\author[inst2]{Erik Cambria\corref{cor1}}
\ead{cambria@ntu.edu.sg}
\author[inst2]{Jagath C. Rajapakse}
\ead{asjagath@ntu.edu.sg}
\cortext[cor1]{Corresponding author}

\affiliation[inst2]{organization={School of Computer Science and Engineering, Nanyang Technological University},
            country={Singapore}}

\begin{abstract}

Entities involve important concepts with concrete meanings and play important roles in numerous linguistic tasks. Entities have different forms in different linguistic tasks and researchers treat those different forms as different concepts. In this paper, we are curious to know whether there are some common characteristics that connect those different forms of entities. Specifically, we investigate the underlying distributions of entities from different types and different languages, trying to figure out some common characteristics behind those diverse entities. After analyzing twelve datasets about different types of entities and eighteen datasets about entities in different languages, we find that while these entities are dramatically diverse from each other in many aspects, their length-frequencies can be well characterized by a family of Marshall-Olkin power-law (MOPL) distributions. We conduct experiments on those thirty datasets about entities in different types and different languages, and experimental results demonstrate that MOPL models characterize the length-frequencies of entities much better than two state-of-the-art power-law models and an alternative log-normal model. Experimental results also demonstrate that MOPL models are scalable to the length-frequency of entities in large-scale real-world datasets.

\end{abstract}


\end{frontmatter}

\section{Introduction}\label{sec:introduction}

\citet{Estoup1916} and~\citet{Zipf1936, Zipf1949} found a very long time ago that the rank-frequency of words in natural languages follows a family of power-law distributions. During his exploration, Zipf also found that the meaning-frequency of words follows power-law distributions as well. The rank-frequency distribution of words is later credited as Zipf's law and provides a direction to understand the use of languages in our communicative system. Zipf's law has been observed in many languages~\citep{Zipf1949, BernatRicard2010} and has attracted tremendous attention of researchers from diverse areas for more than eighty years~\citep{Piantadosi2014}. The Zipf distribution has a linear behavior in the log-log scale and is widely used to model phenomena such as word frequencies, city sizes, income distribution, and network structures. However, the Zipf distribution may not fit well the probabilities of the first positive integer numbers, which are often observed to be higher or lower than expected by the linear model. 

Besides the rank-frequency and meaning-frequency of words, Zipf also analyzed word length, sentence length, and phonemes~\citep{Zipf1949}. Although Zipf explained the use of these three language units under the same principle of least effort as he explained word frequency and word meaning in a qualitative way, unfortunately, extensive studies have demonstrated that the frequencies of these three language units do not follow a power-law distribution, but follow variants of Poisson distributions, lognormal distributions, or gamma distributions~\citep{Williams1940, Fucks1955, Fucks1956, Wake1957, MillerEtal1958Length, Williams1975Mendenhall, GrotjahnAltmann1993, WimmerEtal1994, Best1996, GigurdEtal2004}. 

In the last two decades, the field of natural language processing and related areas have constructed numerous datasets for diverse linguistic tasks~\citep{FSNLP1999, SLP2008, SLP2020}. Those datasets provide us opportunities to analyze some other forms of languages, among which entity is an important one. An entity is a real-world object, such as persons, locations, and organizations~\citep{MUC7, CoNLL2003}. Entities generally involve important concepts with concrete meanings and usually act as (part of) the subject or the object or even both in a sentence. For example, in the sentence ``Michael Jordan could be an NBA player, or a professor of University of California, Berkeley,'' the entity ``Michael Jordan'' acts as the subject while other two entities ``NBA'' and ``University of California, Berkeley'' are parts of the object. Because of its importance in language, entities have been extensively studied and are involved in diverse linguistic tasks, such as named entity recognition~\citep{MUC7, CoNLL2003} and entity linking~\citep{JiAndGrishman2011, LingEtal2015}. To the best of our knowledge, however, there is no existing literature that investigates the underlying distribution(s) of entities which may provide a better understanding on language use and provide insights into designing effective and efficient algorithms for entity-related linguistic tasks.

\begin{table}[t]
\small
\centering
\caption{Some examples of entities in English and their corresponding entity lengths ($l$). Symbols and punctuations in entities are taken into account during the calculation.}
\label{tb:length}
\begin{tabular}{p{0.75\columnwidth}c}
\toprule
\textbf{Entity}								&	Entity Length ($l$)	\\
\midrule
NBA										&	1	\\
Michael Jordan								&	2	\\
United Arab Emirates							&	3	\\
University of California , Berkeley				&	5	\\
10:00 p.m. on August 20 , 1940					&	7	\\
human cytomegalovirus ( HCMV ) major immediate	&	7	\\
\bottomrule
\end{tabular}
\end{table}

In this paper, we fill in this gap and conduct a thorough investigation on the length-frequency distributions of entities in different types and different languages. We aim to fit the length-frequency of entities with a uniform model or a family of models. Entity length is defined by the number of words in an entity. Entity length is an important feature of natural language processing that reflects the complexity and structure of texts. Table~\ref{tb:length} presents some examples of entities and their corresponding lengths. After a careful exploration, we find that the length-frequency of entities cannot be well characterized by pure power-law models, but can be well characterized by the Marshall-Olkin power-law (MOPL) models that are developed by~\citet{perez2013marshall}. MOPL models are a family of generalized models of power-law models. Compared with pure power-law models, MOFL models have more flexibility to adjust the probabilities of the first few data points while keeping the linearity of the remaining probabilities. 

Specifically, we collect twelve datasets about different types of entities (e.g., named entities and time expressions) and eighteen datasets about entities in different languages (e.g., English and French). Those datasets are dramatically diverse from each other in terms of their sources, domains, text genres, generated time, corpus sizes, and entity types, and those languages have significant differences in terms of their phonetic systems and spelling systems (see Section~\ref{ssec:dataset} for details). However, we find that the length of these diverse entities demonstrates some similar characteristics, and the length-frequency distributions of these diverse entities can be well characterized by a family of MOPL models.

To evaluate the quality of MOPL models fitting to the length-frequency of diverse entities, we use the Kolmogorov-Smirnov (KS) test~\citep{smirnov1948table, stephens1974edf} and define an average-error metric to evaluate the goodness-of-fit of the MOFL models and compare the fitting results with two state-of-the-art power-law models, namely CSN2009~\citep{ClausetEtal2009} and $LS_{avg}$~\citep{zhong2022least}, and an alternative log-normal model. We conduct experiments on thirty datasets about entities in different types and different languages, and experimental results demonstrate that MOPL models well characterize the length-frequency distributions of diverse entities, and the fitting results of MOPL are much better than the ones of the three compared models. Specifically, MOPL achieves much better results in the KS test and average-error metric than the three compared models. Experimental results also demonstrate that MOPL models fit the length-frequency of entities in an individual dataset less than one minute, which is comparable with the most efficient model $LS_{avg}$ and much better than the CSN2009 model. This indicates that MOPL models are more suitable to characterize the length-frequency of diverse entities than the three compared models and that MOPL models are scalable to entities in large-scale real-world datasets.\ftext{Source codes and datasets are available at \url{https://github.com/xszhong/MOPL}.}

To summarize, we mainly make in this paper the following contributions.
\begin{itemize}[noitemsep, nolistsep]
\item We investigate the underlying distributions of diverse entities, finding that the length-frequency of entities in different types and languages can be characterized by MOPL models. Our finding adds a piece of stable knowledge to the filed of language and provides insights for entity-related linguistic tasks.
\item We demonstrate the superiority of MOPL models against two state-of-the-art power-law models and a log-normal model in terms of fitting to the length-frequency of diverse entities in different types and languages. 
\item Experiments demonstrate that MOPL is scalable to large-scale real-world datasets without linearly nor exponentially increasing the runtime when the number of entities increases.
\end{itemize}

The remaining of this paper is organized as follows. Section~\ref{sec:literature} reviews the literature about power-law distributions in languages. Section~\ref{sec:method} introduces the MOPL models that we use to characterize the length-frequency of divers entities. Section~\ref{sec:experiment} reports experimental results and computational efficiency of MOPL models and compared models fitting to the length-frequency distributions of entities in different types and different languages. Section~\ref{sec:discussion} discusses possible implications and limitations of this paper while Section~\ref{sec:conclusion} draws the conclusion.

\section{Related Works}\label{sec:literature}

While power-law distributions have been observed to appear in numerous natural systems and societal systems~\citep{PowerLaw2005, ClausetEtal2009}, in this paper, we are concerned with power-law distributions in languages. Following we review related works about the power-law distributions in languages and about the length-frequency distributions of words and sentences, and discuss the connection and differences between these related works and our work.

\subsection{Power-Law Distributions in Languages}\label{ssec:zipflaw}

The most famous power-law distribution in languages is the one in the rank-frequency of words. This linguistic phenomenon was originally discovered by Jean-Baptiste Estoup~\citep{Estoup1916} and then further explored by George K. Zipf~\citep{Zipf1936,Zipf1949}; such linguistic phenomenon is later credited as Zipf's law. Zipf's law reveals that the $r$-th most frequently occurring word in a corpus has the frequency defined by $f(r) \propto r^{-z}$, where $r$ denotes the frequency rank of a word in the corpus and $f(r)$ denotes its frequency. The Zipf's law has been observed in many languages \citep{Zipf1949, Li2002, BernatRicard2010, Piantadosi2014}, and the scaling exponent $z$ is observed to be close to 1. During his exploration, Zipf found as well that the meaning-frequency of words in a corpus also follows a family of power-law distributions. 

Besides real languages, researchers have also explored randomly generated texts and genetic regulatory networks~\citep{pratap2019stability, anbalagan2021razumikhin, pratap2022further}. \citet{Miller1957, Miller1965} and~\citet{Li1992} found that the rank-frequency of random texts also follows power-law distributions. \citet{MaloneMaher2012} and~\citet{WangEtal2017Zipf} found that the rank-frequency of user passwords from different websites can be characterized by power-law distributions.

We now discover another form of human languages, namely entities, whose length-frequency distributions can be characterized by the Marshall-Olkin extended power-law distributions. There are significant differences between power-law distributions in the length-frequency of entities and in the rank-frequency of words. Firstly, the meanings and functions of words and of entities in a sentence are different. In the rank-frequency of words, those most frequent words are always auxiliary words without concrete meanings (random texts and user passwords have no concrete meanings as well), while entities generally involve important concepts with concrete meanings and play important roles in a sentence, such as the subject and the object.

Secondly, the numbers of their data points are different. In the rank-frequency of words, an $r$-rank word appears as a data point, while in the length-frequency of entities, all the $l$-length entities composite a data point. So the number of data points in the rank-frequency of words is as large as the size of vocabulary in a corpus, while the number of data points in the length-frequency of entities is generally less than 100, and our analysis shows that, in about 93.3\% of datasets (28 out of 30), the longest entity contains no more than 100 words (see Table~\ref{tb:dataset-different-type} and~\ref{tb:dataset-different-language}). 

Thirdly, the scaling exponents of these two kinds of power-law distributions are different. The scaling exponents in the rank-frequency of words are observed to approximate to 1, indicating that these power-law distributions do not have theoretical  means nor finite variances. By contrast, the exponents in the length-frequency of entities are greater than 2, theoretically indicating well-defined means in all these power-law distributions; and in real-world datasets, these power-law distributions have finite means and variances.

\subsection{Length-Frequency Distributions of Words and Sentences}\label{ssec:wordlength}

A line of researches that is somewhat related to our work is about the length distributions of words and sentences. According to a review article by~\citet{GrotjahnAltmann1993},~\citet{Fucks1955,Fucks1956} first theoretically and experimentally demonstrated that the length-frequency of words in a corpus follows a family of Poisson distributions. This linguistic phenomenon has been observed in more than 32 languages~\citep{Best1996}. On the other hand,~\citet{Williams1940} and \citet{Wake1957} observed that the length-frequency of sentences in different languages can be characterized by a family of log-normal distributions. \citet{GigurdEtal2004} observed that the length-frequencies of words and sentences from English, Swedish, and German corpora can be characterized by variants of log-normal distributions or gamma distributions.

Unlike the length-frequency of words and sentences that can be characterized by variants of Poisson distributions, log-normal distributions, or gamma distributions, we find from experiments on datasets about entities in different types and different languages that the length-frequency of entities cannot be characterized by Poisson distributions nor log-normal distributions but are well characterized by a family of Marshall-Olkin power-law (MOPL) distributions. Moreover, our extensive experiments demonstrate that MOPL models characterize the length-frequency of entities much better than two state-of-the-art power-law models and one alternative log-normal model and that MOPL models are scalable to the length-frequency of entities in large-scale real-world datasets.

\section{Methodology}\label{sec:method}

We first briefly introduce the discrete power-law distributions and then detail the Marshall-Olkin power-law (MOPL) models that we use to characterize the length-frequency distributions of entities in different types and different languages. After that we introduce the Kolmogorov-Smirnov (KS) test~\citep{smirnov1948table, stephens1974edf} and the average-error metric that are used to evaluate the goodness-of-fit.

\subsection{Discrete Power-Law Distribution}\label{ssec:discrete-powerlaw}

The discrete power-law distribution is given a special case of power-law distributions with discrete values. It is defined by Eq.~(\ref{eq:power-law}): 
\begin{equation}
\label{eq:power-law}
P(X=x) = \frac{x^{-\alpha}}{\zeta(\alpha)}
\end{equation}
where $x \in N^+$, $\alpha>0$ is the scaling exponent, and $\zeta(\alpha) = \Sigma_{k=1}^{\infty} k^{-\alpha}$ is the Riemann Zeta function.

Eq.~(\ref{eq:power-law}) can be written as Eq.~(\ref{eq:log-power-law}), which demonstrates the linear behavior in the log-log scale:
\begin{equation}
\label{eq:log-power-law}
\log P(X=k) = -\alpha \log x - \log\zeta(\alpha)
\end{equation}

The survival function (SF) of the power-law distribution is given by Eq.~(\ref{eq:sf_power-law}):
\begin{equation}
\label{eq:sf_power-law}
\overline{F}(X) = P(X > x) =  \frac{\zeta(\alpha, x+1)}{\zeta(\alpha)}
\end{equation}
where  $\zeta(\alpha, x) = \Sigma_{k=x}^{\infty} k^{-\alpha}$ is the Hurwitz zeta function.

\subsection{Marshall-Olkin Power-Law Distribution}\label{ssec:MOPLfit}

\citet{perez2013marshall} explore a new form of power-law distributions by extending the original power-law function through the Marshall-Olkin transformation. They extend the original power-law function to a more general function called Marshall-Olkin power-law distribution. This function have two parameters, $\alpha$ and $\beta$, and its survival function (SF) is given as below:
\begin{equation}
\label{eq:MOPL_SF}
P(X>x) = \overline G(x; \alpha, \beta) = \frac{\beta\overline F(X)}{1-\overline\beta\overline F(X)} = \frac{\beta\zeta(\alpha, x+1)}{\zeta(\alpha)-\overline\beta\zeta(\alpha + 1)}
\end{equation}
where $\beta>0$, $\alpha > 1$ and $\overline\beta = 1 - \beta$. 

The probability mass function (PMF) can be computed through Eq.~(\ref{eq:MOPL_pmf}):
\begin{equation}
\begin{aligned}
P(X&=x) = \overline G(x-1; \alpha, \beta) - \overline G(x; \alpha, \beta)\\
&= \frac{x^{-\alpha}\beta\zeta(\alpha)}{[\zeta(\alpha) - \overline{\beta}\zeta(\alpha, x)][\zeta(\alpha) - \overline(\beta)\zeta(\alpha, x+1)]}
\end{aligned}
\label{eq:MOPL_pmf}
\end{equation}
where $x\in N^+$ and $\zeta(\alpha, x)=\Sigma_{k=x+1}^{\infty} k^{-\alpha}$ stands for the Hurwitz Zeta function.

The Marshall-Olkin power-law (MOPL) distributions are a generalization of power-law distributions and overcome some limitations of pure power-law distributions by introducing a parameter. Such parameter allows for more flexibility in adjusting the probabilities of small values while keeping the linearity in tails. The MOPL models are capable of fitting the concave and convex issues encountered in realistic situations, and have been applied to characterize various data such as music compositions and web page visits~\citep{perez2013marshall}.

In this paper, we use the MOPL models to characterize the length-frequency distributions of entities in different types and different languages.

\subsection{Kolmogorov-Smirnov Test}\label{ssec:kstest}

Like many previous researches~\citep{ClausetEtal2009, HanelEtal2017, WangEtal2017Zipf, GerlachAltmann2019, ArticoEtal2020, NettasingheKrishnamurthy2021KDD, zhong2022least}, we employ the Kolmogorov-Smirnov (KS) test~\citep{smirnov1948table, stephens1974edf} to examine the goodness-of-fit. The KS statistic ($D_n$) quantifies the distance between the cumulative distribution function (CDF) of a set of data points ($F_n(l)$) and the CDF of a theoretic distribution ($F(l)$), as defined by Eq.~(\ref{eq:kstest}):

\begin{equation}
\label{eq:kstest}
D_n=\sup_l{|F_n(l)-F(l)|}
\end{equation}
where $\sup_l$ is the supremum of the set of distances. The KS statistic $D_n\in [0,1]$ is the maximal distance between the two CDF curves $F_n(l)$ and $F(l)$. The smaller the $D_n$ value is, the better the theoretic distribution fits the data points.

The KS test can also be used to examine whether two underlying distributions are significantly different. In such case, the two-sample KS statistic ($D_{n,m}$) is defined by Eq.~(\ref{eq:two-kstest}):

\begin{equation}
\label{eq:two-kstest}
D_{n,m}=\sup_l{|F_n(l)-F_m(l)|}
\end{equation}
where $F_n(l)$ and $F_m(l)$ are the CDF curves of two sets of data points.

In the KS test, the null hypothesis ($H_0$) is that the data points are drawn from a theoretic distribution, where the theoretic distribution can be any parametric distribution, such as zipf distribution, normal distribution, power law distribution, and lognormal distribution; the alternative ($H_1$) is that the data points are not drawn from the theoretic distribution. A larger $p$-value suggests that it is safer to draw a conclusion that these data points are not significantly different from the hypothesized distribution. In two-sample KS test, the null hypothesis (${H}'_0$) is that the two sets of data points are drawn from the same underlying distribution, while the alternative (${H}'_1$) is that they are not from the same distribution. Similarly, a larger $p$-value suggests that it is safer to draw a conclusion that the two sets of data points are drawn from the same underlying distribution. 

\subsection{Average Error}\label{ssec:avgerr}

Besides the KS test, we also define a metric called average error to examine the goodness-of-fit. The average error is defined by Eq.~(\ref{eq:avgerr}):

\begin{equation}
\label{eq:avgerr}
E_{avg}=\frac{1}{N} \sum_{x_i} \frac{\left|p_N\left(x_i\right)-p\left(x_i\right)\right|}{\sqrt{p_N\left(x_i\right) \cdot p\left(x_i\right)}}
\end{equation}
where $p_N(x)$ and $p(x)$ are the probability density functions (PDF) of the raw data and the hypothesized data. $N=\mid\left\{\left(x_i, p_N\left(x_i\right)\right\} \mid\right.$ stands for the number of data points. Defining the average-error metric by Eq.~(\ref{eq:avgerr}) is to remove the impact of different sample sizes. For different models fitting to the same dataset, the smaller the model achieves the $E_{avg}$, the better the model fits the dataset.

\section{Experiments}\label{sec:experiment}

We fit Marshall-Olkin power-law (MOPL) models to twelve datasets about different types of entities and eighteen datasets about entities in different languages and compare the fitting results of MOPL with two state-of-the-art models, namely CSN2009~\citep{ClausetEtal2009} and $LS_{avg}$~\citep{zhong2022least}, and an alternative log-normal model.

\subsection{Datasets}\label{ssec:dataset}

The datasets we use in this paper mainly involve two kinds: (1) entities in different types and (2) entities in different languages. Most of these datasets contain annotated entities while some contain automatically annotated entities. We collect from both their training and test sets of these datasets for their entities.

\subsubsection{Entities in Different Types}\label{sssec:different-type}

This kind of datasets contains twelve datasets regarding different types of entities collected from dramatically diverse sources, including general named entities~\citep{MUC6, MUC7, CoNLL2003}, entity mentions~\citep{LingAndWeld2012, OntoNotes2013}, time expressions~\citep{TimeML2003, TimeBank2003, Zhong2023Time}, aspect terms~\citep{Liu2012, PontikiEtal2014}, literary entities~\citep{Litbank2019}, defense entities, informal entities~\citep{RitterEtal2011, BTC16}, and domain-specific entities~\citep{FukudaEtal1998, TakeuchiAndCollier2005} that are well studied in the field of natural language processing and related areas. In this paper, we use the term of ``\textit{entity}'' to broadly represent these diverse concepts, and these specific concepts are treated as \textit{different types of entities}. In a specific type of entities, researchers may also assign some pre-defined labels (e.g., PERSON, LOCATION, and ORGANIZATION) to these entities. We use ``different types of entities'' or ``entity types'' to represent the above general named entities, time expressions, aspect terms, etc., while use ``different categories of entities'' or ``entity categories'' to represent these pre-defined labels. In our analysis, we are concerned with ``different types of entities'' and do not care much about ``different categories of entities.'' Because each type of entities may also contain different categories/labels and can reveal general habits of our humans in using language, while a certain category of entities reveal only our specific/narrow habit(s). In this paper, we care more about those general habits and principles than specific/narrow one(s). Since English is the most studied language in natural language processing and related areas, we analyze these different types of entities in English. 

The twelve datasets are (1) ABSA~\citep{PontikiEtal2014, PontikiEtal2015}, (2) ACE04~\citep{ACE2004}, (3) BBN~\citep{BBN2005}, (4) BioMed~\citep{CrichtonEtal2017}, (5) CoNLL03~\citep{CoNLL2003}, (6) COVID19~\citep{COVID19NER}, (7) LitBank~\citep{Litbank2019}, (8) OntoNotes5~\citep{OntoNotes2013}, (9) Re3d, (10) TimeExp~\citep{TimeBank2003, WikiWars2010, TempEval-3, SynTime2017, TOMN2018}, (11) Twitter~\citep{WNUT16, BTC16}, (12) WikiAnchor~\citep{LingAndWeld2012}. They are briefly described below in alphabetical order.

\textbf{ABSA} contains two corpora that are used in SemEval-2014~\citep{PontikiEtal2014} and SemEval-2015~\citep{PontikiEtal2015} for aspect-based sentiment analysis. While the two corpora have several language units for different tasks, we are concerned with aspect terms and collect these aspect terms for the analysis of their length-frequency distribution.

\textbf{ACE04} is a benchmark dataset used for the 2004 Automatic Content Extraction (ACE) technology evaluation~\citep{ACE2004}. It consists of various types of data collected from different sources (e.g., newswire and broadcast news) for the analysis of entities and relations in three languages: Arabic, Chinese, and English. We use its English entities for the analysis of different types of entities, while use its Arabic entities for the analysis of entities in different languages.

\textbf{BBN} consists of Wall Street Journal articles for pronoun co-reference and entity analysis~\citep{BBN2005}. It includes 28 entity categories in total. We collect all of its entities for analysis, without considering its entity categories.

\textbf{BioMed} contains fourteen corpora that are developed for the analysis of biomedical entities. \citet{CrichtonEtal2017} collect the fourteen corpora and we can get these corpora from their paper for the biomedical entities.

\textbf{CoNLL03} is a benchmark dataset with 1,393 news articles derived from the Reuters RCV1 Corpus, which is collected between the period of August 1996 and August 1997~\citep{CoNLL2003}. We collect its entities without entity categories for the analysis of the length-frequency distribution.

\textbf{COVID19} is a newly constructed dataset for the analysis of entities related to the recent COVID-19 pandemic~\citep{COVID19NER}. We collect and analyze its entities for the length-frequency analysis.

\textbf{LitBank} is a dataset collected from 100 different English-language literary articles across over a long period of time and it is developed for the analysis of literary entities~\citep{Litbank2019}.

\textbf{OntoNotes5} is a large-scale dataset collected from different sources (e.g., news articles, newswire and web data) over a long period of time for the comprehensive analyses of syntax, co-reference, proposition, word sense, and named entities in three languages (i.e., English, Chinese, and Arabic)~\citep{OntoNotes2013}. In this paper we are concerned with its entities in English for analysis.

\textbf{Re3d}~\furl{https://github.com/dstl/re3d} is a dataset with various documents relevant to the conflict in Syria and Iraq. The dataset is constructed for the analysis of entity and relation extraction in the domain of defense and security. We collect its entities for analysis.

\begin{table}
\small
\centering
\caption{Statistics of datasets about entities in different types. Entity length $l$ is defined by the number of words in an entity.}
\label{tb:dataset-different-type}
\begin{tabular}{@{}llrrcr@{}}
\toprule
\textbf{Dataset}	& \textbf{Entity Type}	&  \textbf{Num of Entities}   & \textbf{Max $l$}	&   \textbf{Average $l$}	&	\textbf{StdDev. $l$}\\
\midrule
ABSA	&	aspect terms	&	9,979		&	21  	&	1.45	&	0.89	\\
ACE04	&	named entities	&	29,949	&	57  	&	2.43	&	9.29	\\
BBN		&	named entities	&	98,427	&	15	&	1.26	&	0.36	\\
BioMed	&	biomedical entities&	450,729	&	86  	&	1.80	&	4.05	\\
CoNLL03	&	named entities	&	35,087	&	14	&	1.45	&	0.48	\\
COVID19	&	pandemic entities&	10,260,797&	117	&	1.27	& 	0.63	\\
LitBank      &	literary entities  	&	13,912	&	129 	&	2.93	&	19.66	\\
OntoNotes5&	named entities	&	155,413	&	28 	&	1.85	&	1.58	\\
Re3d 	&	defense entities	&	3,394		&	20	&	2.32	&	3.20	\\
TimeExp	&	time expressions	&	18,484	&	22  	&	1.80	&	1.31	\\
Twitter	&	informal entities	&	20,515	&	14  	&	1.39	&	0.71	\\
WikiAnchor&	anchor text		&	2,690,849	&	49 	&	2.10	&	3.09	\\
\bottomrule
\end{tabular}
\end{table}

\begin{table}[t]
\small
\centering
\caption{Statistics of entities in different languages}
\label{tb:dataset-different-language}
\begin{tabular}{llrccr}
\toprule
\textbf{Language}	&	\textbf{Entity Type}	& \textbf{Num of Entities}	&    \textbf{Max $l$}   &  \textbf{Average $l$}	&	\textbf{StdDev. $l$}\\
\midrule
Afrikaans	&	named entities	&	13,947	&	27	&	1.86	&	1.87	\\
Arabic	&	named entities	&	44,284	&	41	&	2.15	&	6.06\\
Basque	&	named entities	&	4,748		&	20	&	1.47	&	0.62	\\
Bokmal	&	named entities	&	13,950	&	15	&	1.10	&	0.19	\\
Croatian	&	named entities	&	21,105,675	&	11	&	1.95	&	2.37	\\
Czech	&	named entities	&	62,867	&	9	&	1.53	&	0.79	\\
France	&	named entities	&	9,836		&	17	&	1.41	&	0.75	\\
German	&	named entities	&	12,778	&	34	&	1.53	&	0.91	\\
Italian	&	named entities	&	1,071,045	&	41	&	2.35	&	2.37	\\
Netherland	&	named entities	&	7,102		&	9	&	1.42	&	0.99	\\
Nynorsk	&	named entities	&	12,726	&	10	&	1.13	&	0.25	\\
Polish	&	named entities	&	12,038,419	&	13	&	1.86	&	1.16	\\
Romanian	&	named entities	&	153,226	&	30	&	1.77	&	1.94	\\
Russian	&	named entities	&	3,152,930	&	12	&	1.70	&	1.16	\\
Samnorsk	&	named entities	&	29,407	&	15	&	1.11	&	0.22	\\
Slovak	&	named entities	&	136435	&	11	&	1.72	&	1.44	\\
Slovene	&	named entities	&	13,055,756	&	8	&	2.07	&	2.03	\\
Ukrainian	&	named entities	&	18,347,492	&	14	&	2.23	&	2.31	\\
\bottomrule
\end{tabular}
\end{table}

\textbf{TimeExp} consists of three corpora that are developed for the analysis of time expressions~\citep{SynTime2017, TOMN2018, UGTO2020}. These corpora include TempEval-3 (including TimeBank~\citep{TimeBank2003}, TE3-Silver, AQUAINT, and the Platinum corpus)~\citep{TempEval-3}, WikiWars~\citep{WikiWars2010}, and Tweets~\citep{SynTime2017}.

\textbf{Twitter} consists of two corpora whose text is collected from Twitter: WNUT16 \citep{WNUT16} and Broad Twitter Corpus~\citep{BTC16}. These two corpora are developed for the analysis of entities in informal text.

\textbf{WikiAnchor} treats the anchor text (\ie the text in the hyperlinks) from Wikipedia (the 20110513 version) as entity mentions~\citep{LingAndWeld2012}. We collect these entity mentions (i.e., anchor text) for length-frequency analysis.

For each of these datasets that contain two or more corpora (i.e., ABSA, BioMed, TimeExp, and Twitter), we simply merge all the entities from the whole corpora. Note again that we collect from these datasets only their entities for the analysis of length-frequency distribution; we do not care about their entity categories (or pre-defined labels).

Table~\ref{tb:dataset-different-type} reports the entity types and statistics of the twelve datasets. As mentioned in Section~\ref{ssec:MOPLfit}, the entity length $l$ is defined by the number of words in an entity. Table~\ref{tb:dataset-different-type} shows that the numbers of entities in the twelve datasets are diverse dramatically, ranging from 3,394 (Re3d) to 10,260,797 (COVID19); and the maximal lengths and standard deviations of these entities are also diverse: the maximal lengths are varied from 14 to 129 and the standard deviations are varied from 0.36 to 19.66, respectively. However, the average lengths of these entities are comparable and range around 2 (only from 1.26 to 2.93). This indicates that the average length is a common characteristic among these diverse entities.

\subsubsection{Entities in Different Languages}\label{sssec:different-language}

This kind of datasets contains named entities in eighteen different languages. These datasets are collected from 2004 Automatic Content Extraction (ACE) evaluation~\citep{ACE2004}, European Newspapers\furl{https://github.com/EuropeanaNewspapers/ner-corpora}, NCHLT Afrikaans Named Entity Annotated Corpus\furl{https://repo.sadilar.org/handle/20.500.12185/299}, Basque EIEC (version 1.0)\furl{http://www.ixa.eus/node/4486?language=en}, BSNLP 2017\furl{http://bsnlp-2017.cs.helsinki.fi/shared_task.html}, Italian KIND~\citep{paccosi2021kind}, Norwegian Navnkjenner~\citep{johansen2019ner}, and RONEC~\citep{dumitrescu2019introducing}.

The eighteen languages include (1) Afrikaans, (2) Arabic, (3) Basque, (4) Bokmal, (5) Croatian, (6) Czech, (7) France, (8) German, (9) Italian, (10) Netherland, (11) Nynorsk, (12) Polish, (13) Romanian, (14) Russian, (15) Samnorsk, (16) Slovak, (17) Slovene, and (18) Ukrainian. We do not include English in this kind of datasets because different types of entities are analyzed in English. Table~\ref{tb:dataset-different-language} summarizes the statistics of entities in the eighteen languages. It shows that the numbers of these entities are significantly diverse, ranging from 4,748 (Basque) to 21,105,675 (Croatian). The maximal lengths and standard deviations of these entities in different languages are somewhat diverse but not that dramatical; while the average lengths of these entities are comparable, ranging around 2 (specifically, from 1.10 to 2.35). These statistics are consistent with corresponding ones of different types of entities reported in Table~\ref{tb:dataset-different-type}. This indicates that entities across different types and different languages share some similar characteristics.

\subsection{Compared Methods}\label{ssec:compared-method}

We evaluate the quality of MOPL models in fitting the length-frequency distributions of entities against two state-of-the-art models, namely CSN2009~\citep{ClausetEtal2009} and $LS_{avg}$~\citep{zhong2022least}, and an alternative log-normal model.

\textbf{CSN2009}: \citet{ClausetEtal2009} propose a maximum-likelihood fitting method, which is denoted by CSN2009, that combines with goodness-of-fit tests based on the Kolmogorov-Smirnov statistic to fit power-law distributions to empirical data. CSN2009 estimates the exponent of a power-law model and the minimal value from which the power-law distribution starts. Besides data fitting, CSN2009 also adopts the KS test with likelihood ratios to evaluate the goodness-of-fit of how well a model fits to data. CSN2009 has been the most popular method in the last decade in fitting power-law distributions.

\textbf{$LS_{avg}$}: \citet{zhong2022least} demonstrate through extensive experiments that least-squares methods can accurately fit to power-law distributions. They propose a least-squares method to fit power-law distributions to empirical data and use an average strategy to reduce the impact of noisy data that deviate from the fitted line.

\textbf{LogNormal}: Log-normal distributions are alternative distributions that researchers usually use to fit data when considering power-law distributions. Therefore, besides CSN2009 and $LS_{avg}$, we also compare MOPL models with the log-normal model in terms of fitting the length-frequency of entities.

\begin{table}[t]
\small
\centering
\caption{Fitting results of MOPL and compared models fitting to the length-frequency distributions of entities in different types. $C$ indicates the coverage which is defined by the percentage of data covered by a model. $M_{log}$ denotes logarithmic mean while $V_{log}$ denotes logarithmic variance.}
\label{tb:parameter-different-type}
\begin{tabular}{
@{}p{1.6cm}
c@{\hspace{2mm}}r@{\hspace{2mm}}r
c@{\hspace{2mm}}r
c@{\hspace{1mm}}c@{\hspace{1mm}}r
c@{\hspace{2mm}}c@{\hspace{2mm}}r@{}}
\toprule
\multirow{2}{*}{\textbf{Dataset}} & \multicolumn{3}{c}{\textbf{MOPL}} & \multicolumn{2}{c}{$LS_{avg}$} & \multicolumn{3}{c}{\textbf{CSN2009}} & \multicolumn{3}{c}{\textbf{LogNormal}} \\
\cmidrule(lr){2-4}\cmidrule(lr){5-6}\cmidrule(lr){7-9}\cmidrule(lr){10-12}
    & $\hat\alpha$ & \centering{$\hat\beta$}	&	$C$(\%) & $\hat\alpha$	&	$C$(\%)  & $\hat{\alpha}$	& $\hat{x}_{min}$	&	$C$(\%)  & $M_{log}$ & $V_{log}$	&	$C$(\%)  \\
\midrule
ABSA	&	4.07&5.44&99.82	&	2.34&99.95	&	3.68&2&28.79	&	0.26&0.19&100.00\\
ACE04	&	2.69&2.50&99.54	&	1.61&99.97	&	2.73&4&15.38	&	0.55&0.51&100.00\\
BBN		&	4.74&5.43&99.97	&	3.03&100.00	&	6.77&4&1.23	&	0.16&0.11&100.00\\
BioMed	&	2.84&2.17&99.92	&	2.02&99.99	&	3.36&4&8.53	&	0.36&0.33&100.00\\
CoNLL03	&	5.83&29.48&99.97	&	2.51&100.00	&	5.09&2&36.78	&	0.28&0.15&100.00\\
COVID19	&	3.68&1.94&99.00	&	2.42&99.99	&	4.96&4&2.10	&	0.15&0.13&100.00\\
LitBank	&	3.44&14.98&99.47	&	2.94&99.68	&	2.61&2&70.99	&	0.62&0.41&100.00\\
OntoNotes5&	3.71&3.12&99.90	&	0.73&99.99	&	5.31&5&1.28	&	0.22&0.17&100.00\\
Re3d		&	3.26&8.79&98.70	&	1.12&99.82	&	4.67&6&5.10	&	0.69&0.55&100.00\\
TimeExp	&	4.19&14.15&99.91	&	1.46&100.00	&	5.34&4&8.09	&	0.45&0.26&100.00\\
Twitter	&	4.20&5.21&99.91	&	2.54&99.99	&	3.86&2&26.19	&	0.23&0.16&100.00\\
WikiAnchor&	4.21&23.02&100.00	&	2.55&100.00	&	3.81&3&24.69	&	0.58&0.30&100.00\\
\bottomrule
\end{tabular}
\end{table}

\begin{sidewaystable}[htp]
\small
\centering
\caption{Goodness-of-fit testing results of MOPL and compared models fitting to the length-frequency distributions of entities in different types. $D_{n}$ indicates the KS statistic defined by Eq.~(\ref{eq:kstest}). $E_{avg}$ indicates the average error defined by Eq.~(\ref{eq:avgerr}). $DEC$ indicates the decision to accept or reject the hypothesis $H_0$ that a model well fits the data, based on the $p$-value of the KS test. For each of $D_{n}$ and $E_{avg}$, the best result on each dataset is highlighted in bold.}
\label{tb:result-different-type}
\begin{tabular}{
  @{}l
  p{1.6cm}@{\hspace{2mm}}
  p{0.6cm}
  p{1.1cm}
  p{1.6cm}@{\hspace{2mm}}
  p{0.6cm}
  p{1.1cm}
  p{1.6cm}@{\hspace{2mm}}
  p{0.6cm}
  p{1.1cm}
  p{1.6cm}@{\hspace{2mm}}
  p{0.6cm}
  p{1.1cm}
  @{}
}
\toprule
\multirow{2}{*}{\textbf{Dataset}}	&	\multicolumn{3}{c}{\textbf{MOPL}}	&	\multicolumn{3}{c}{$LS_{avg}$}&\multicolumn{3}{c}{\textbf{CSN2009}}&	\multicolumn{3}{c}{\textbf{LogNormal}}		\\
\cmidrule(lr){2-4}\cmidrule(lr){5-7}\cmidrule(lr){8-10}\cmidrule(lr){11-13}
	&	\centering{$D_n$}	&	$E_{avg}$	&	$DEC$	&	\centering{$D_n$}	&	$E_{avg}$	&	$DEC$	
	&	\centering{$D_n$}	&	$E_{avg}$	&	$DEC$	&	\centering{$D_n$}	&	$E_{avg}$	&	$DEC$\\
\midrule

ABSA&\textbf{1.67E-03}&\textbf{0.18}&\textbf{Accept}&4.17E-01&1.48&Reject&2.63E-02&0.35&Reject&3.97E-02&1.28&Reject\\
ACE04&\textbf{6.15E-03}&\textbf{0.18}&\textbf{Accept}&5.28E-01&1.60&Reject&4.29E-02&0.32&Reject&1.21E-01&1.51&Reject\\
BBN&\textbf{6.51E-04}&0.43&\textbf{Accept}&2.73E-01&1.88&Reject&1.24E-02&\textbf{0.25}&\textbf{Accept}&5.69E-02&4.61&Reject\\
BioMed&\textbf{1.58E-03}&0.62&\textbf{Accept}&6.27E-01&2.61&Reject&9.71E-03&\textbf{0.34}&Reject&1.15E-01&3.28&Reject\\
CoNLL03&\textbf{3.36E-04}&\textbf{0.32}&\textbf{Accept}&3.33E-01&2.34&Reject&4.46E-03&0.36&\textbf{Accept}&6.68E-02&1.11&Reject\\
COVID19&\textbf{7.88E-05}&1.40&\textbf{Accept}&6.25E-01&3.96&Reject&8.69E-03&\textbf{0.66}&Reject&4.97E-02&11.27&Reject\\
LitBank&\textbf{1.73E-03}&0.87&\textbf{Accept}&8.00E-01&3.39&Reject&2.00E-02&\textbf{0.32}&Reject&6.50E-02&0.92&Reject\\
OntoNotes5&\textbf{2.04E-03}&0.51&\textbf{Accept}&3.85E-01&1.60&Reject&1.83E-02&\textbf{0.30}&\textbf{Accept}&5.40E-02&2.66&Reject\\
Re3d&\textbf{1.22E-02}&\textbf{0.28}&\textbf{Accept}&4.62E-01&1.53&Reject&6.02E-02&0.39&\textbf{Accept}&5.64E-02&0.36&Reject\\
TimeExp&\textbf{1.22E-03}&0.37&\textbf{Accept}&5.88E-01&4.57&Reject&1.00E-02&\textbf{0.36}&\textbf{Accept}&3.14E-02&0.72&Reject\\
Twitter&\textbf{1.24E-03}&\textbf{0.21}&\textbf{Accept}&3.33E-01&1.22&Reject&1.92E-02&0.36&Reject&4.02E-02&2.21&Reject\\
WikiAnchor&\textbf{1.63E-04}&0.92&\textbf{Accept}&2.92E-01&1.12&Reject&1.20E-02&\textbf{0.59}&Reject&1.76E-02&4.46&Reject\\
\bottomrule
\end{tabular}
\end{sidewaystable}

\subsection{Implementation Details}\label{ssec:comparison}

For the experiments of data fitting, we use the zipfextR package~\citep{perez2013marshall} in the R programming language to implement our method and apply the codes of CSN2009~\furl{https://aaronclauset.github.io/powerlaws/} and $LS_{avg}$~\furl{https://github.com/xszhong/LSavg} to the datasets. For the KS test, we use the \textit{dgof}~\furl{https://cran.r-project.org/web/packages/dgof/index.html}~\citep{arnold2011nonparametric} and \textit{KSgeneral}~\furl{https://github.com/raymondtsr/ksgeneral}~\citep{JSSv095i10} packages in the R programming language for MOPL, $LS_{avg}$, and the log-normal model, while use CSN2009's KS-test module for CSN2009. In experiments, we find that for the same model on the same dataset, \textit{dgof} and \textit{KSgeneral} achieve the same $D_n$ value (i.e., the KS statistic) but different $p$-values. This suggests that the $D_n$ values are accurate while the $p$-values may not be accurate. In this paper, we use the \textit{dgof} package to report the $D_n$ values and make the final Accept/Reject decisions. All our experiments are conducted on a Dell PowerEdge R740 server with a 96-CPUs processor, 256GB memory, and the CentOS-7 system.

\subsection{Experimental Results}\label{ssec:experiment-result}

\begin{figure}[!htp]
\centering
\subfigure[\label{subfig:ABSA}ABSA]{\includegraphics[width=0.325\columnwidth]{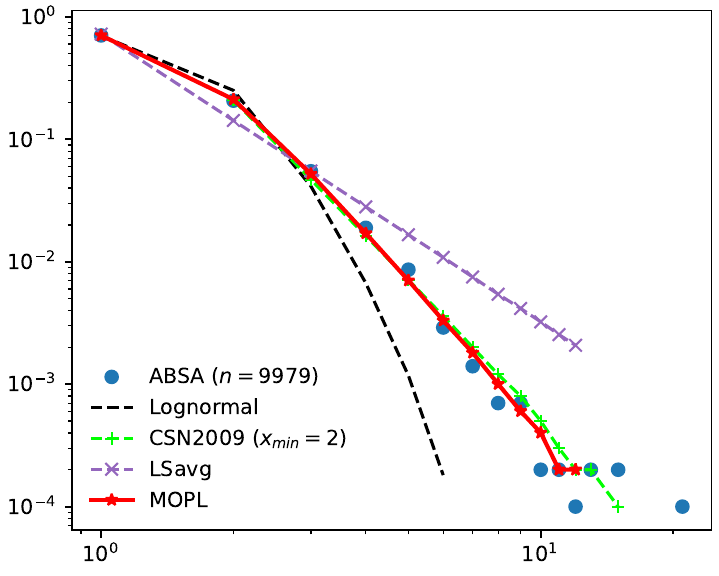}}
\subfigure[\label{subfig:ace04}ACE04]{\includegraphics[width=0.325\columnwidth]{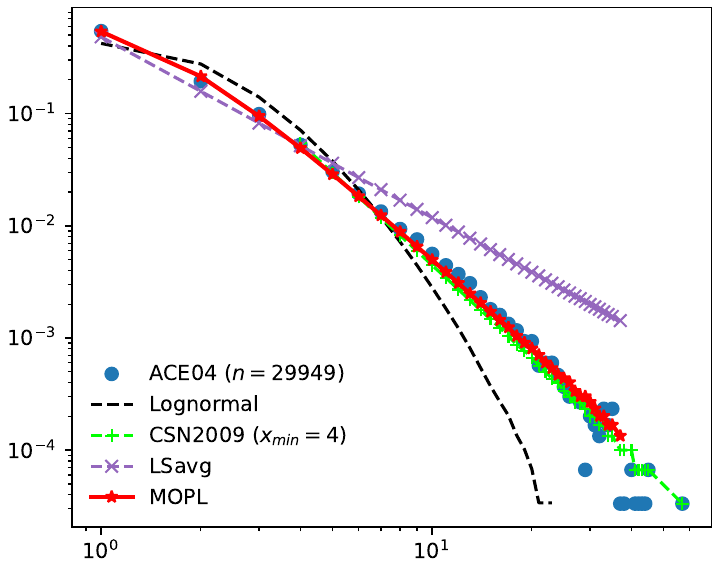}}
\subfigure[\label{subfig:BBN}BBN]{\includegraphics[width=0.325\columnwidth]{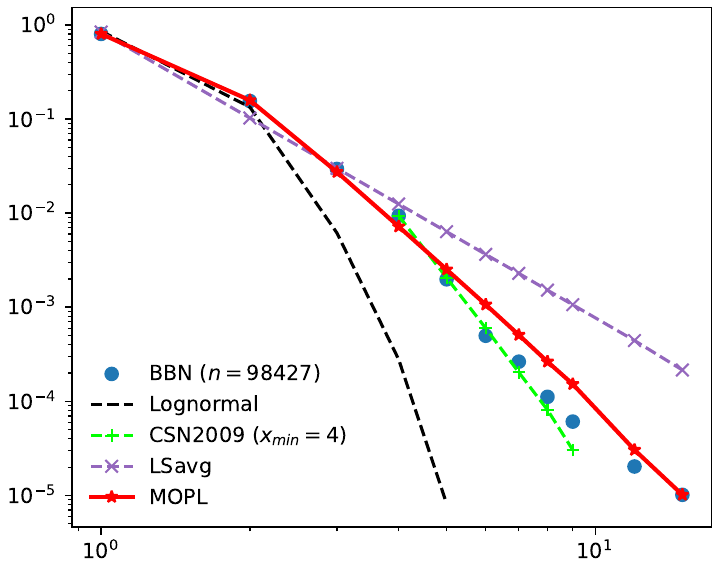}}
\subfigure[\label{subfig:BioMed}BioMed]{\includegraphics[width=0.325\columnwidth]{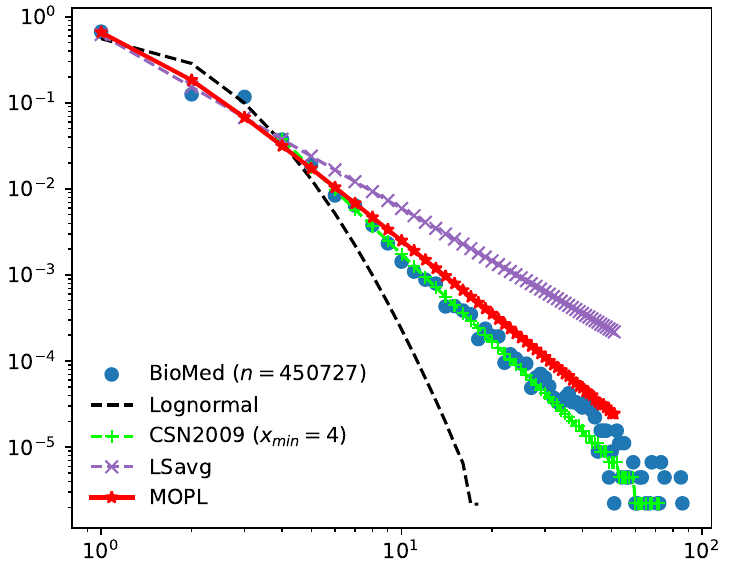}}
\subfigure[\label{subfig:CoNLL03}CoNLL03]{\includegraphics[width=0.325\columnwidth]{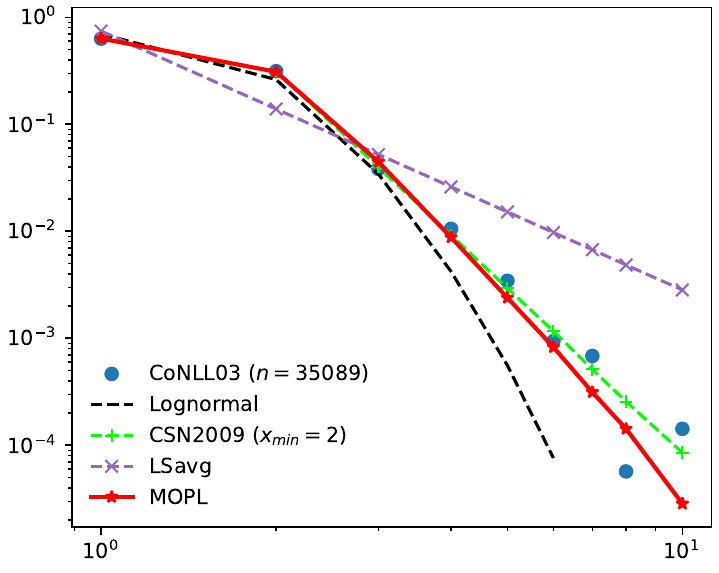}}
\subfigure[\label{subfig:COVID19}COVID19]{\includegraphics[width=0.325\columnwidth]{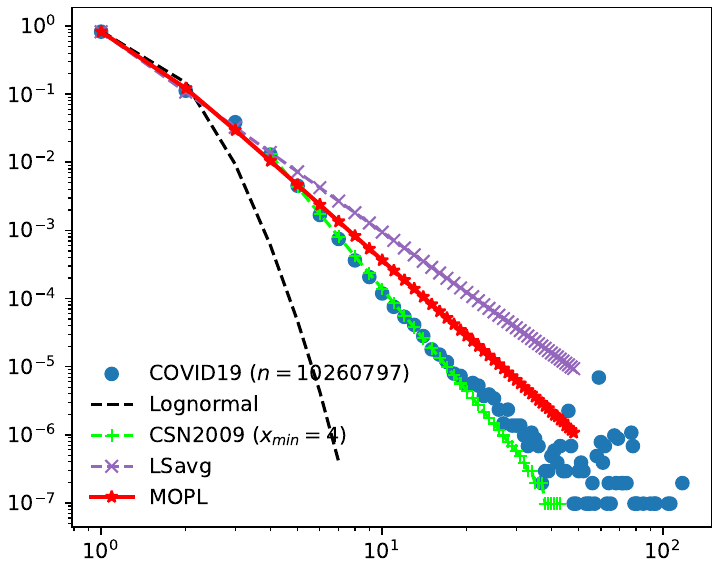}}
\subfigure[\label{subfig:LitBank}LitBank]{\includegraphics[width=0.325\columnwidth]{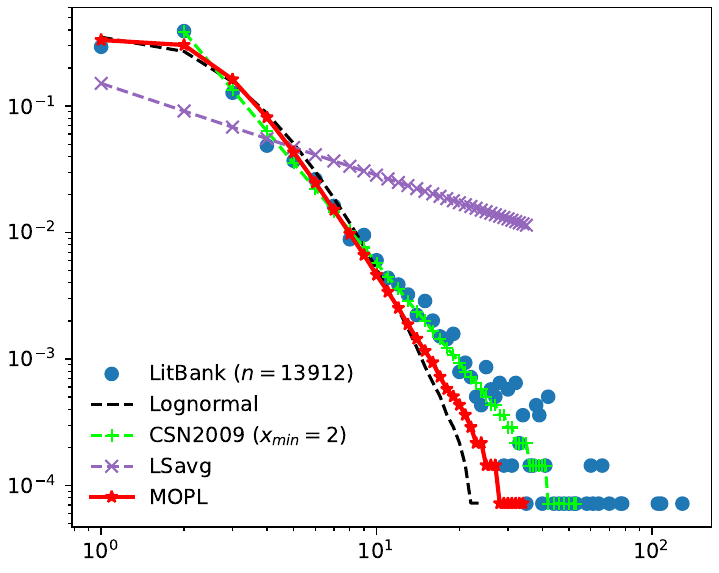}}
\subfigure[\label{subfig:OntoNotes5}OntoNotes5]{\includegraphics[width=0.325\columnwidth]{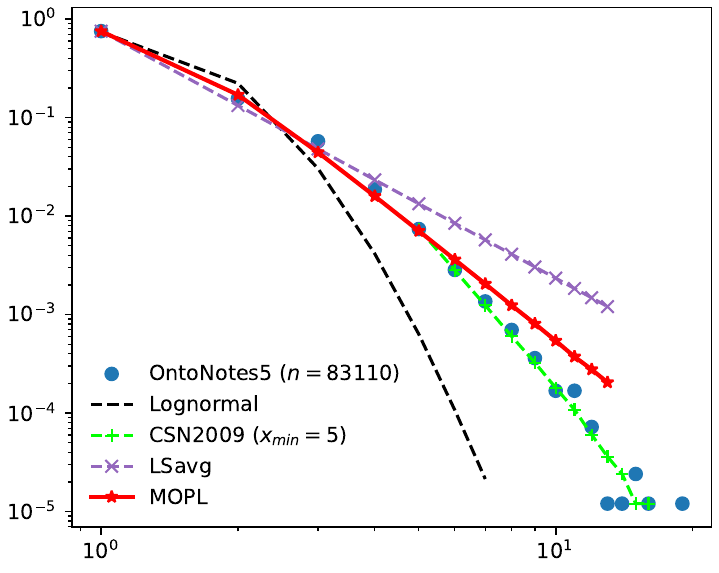}}
\subfigure[\label{subfig:Re3d}Re3d]{\includegraphics[width=0.325\columnwidth]{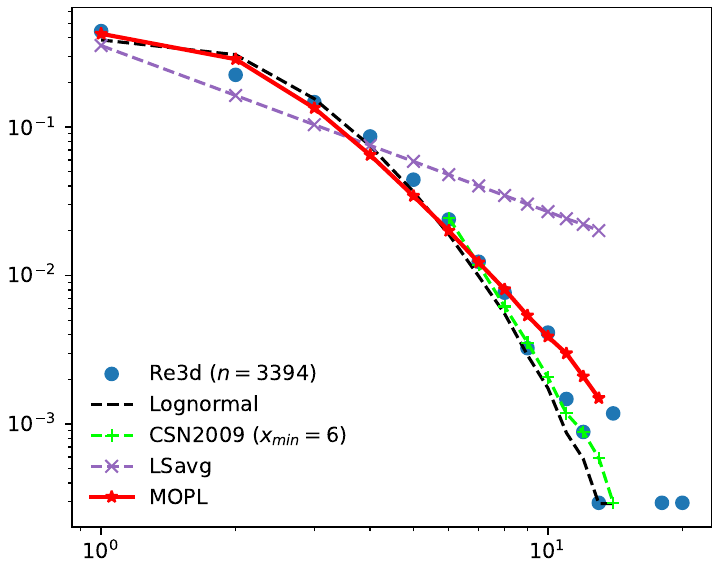}}
\subfigure[\label{subfig:TimeExp}TimeExp]{\includegraphics[width=0.325\columnwidth]{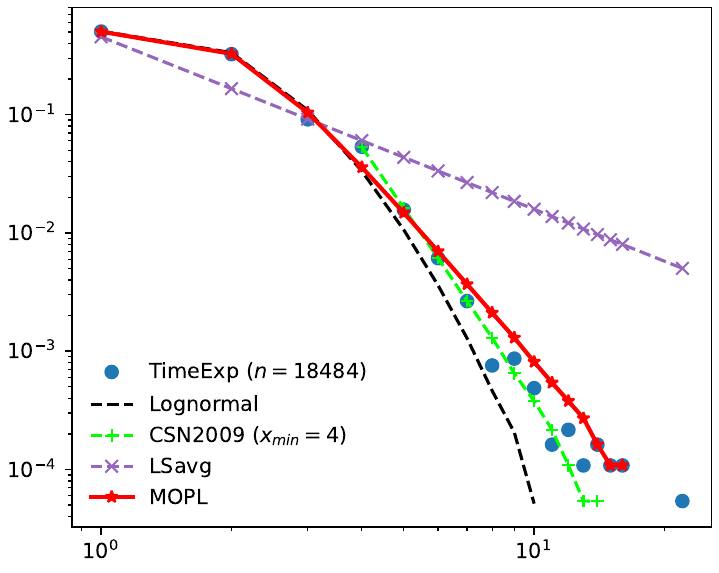}}
\subfigure[\label{subfig:Twitter}Twitter]{\includegraphics[width=0.325\columnwidth]{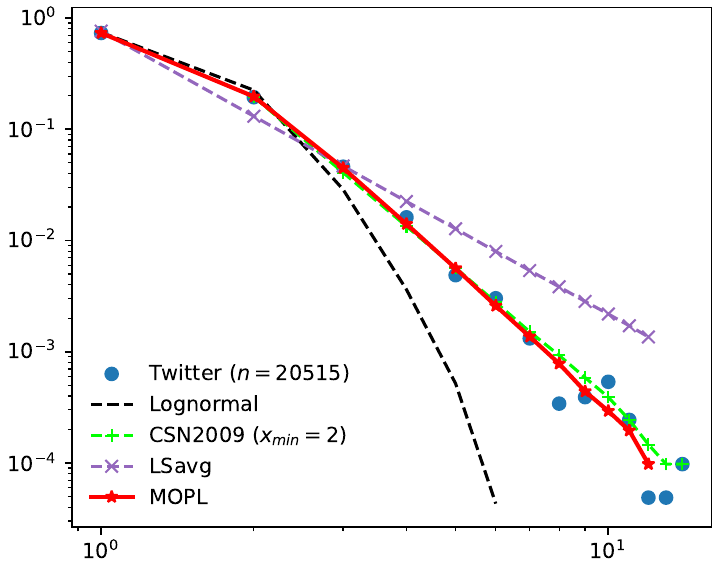}}
\subfigure[\label{subfig:WikiAnchor}WikiAnchor]{\includegraphics[width=0.325\columnwidth]{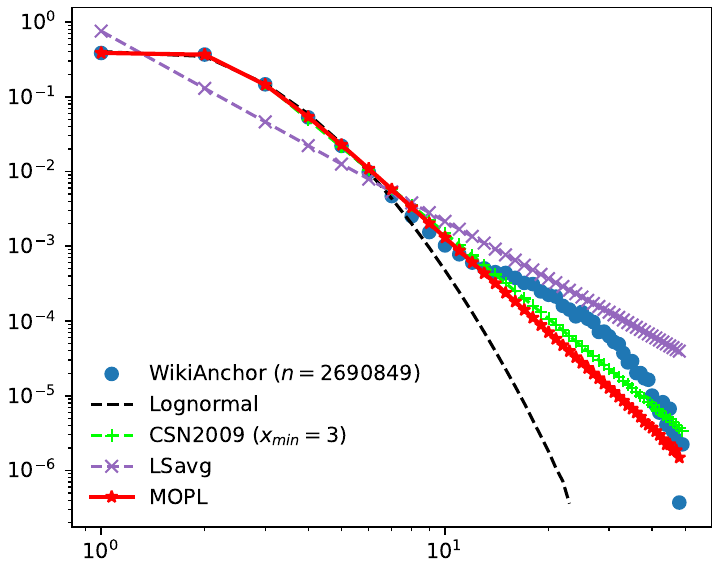}}
\caption{Plots of MOPL and the three compared models fitting to the length-frequency distributions of entities in different types in the twelve datasets. The horizontal axis indicates the entity length ($l$) while the vertical axis indicates the percentage ($p(l)$).}
\label{fig:fitting-type}
\end{figure}

\begin{figure}[!htp]
\subfigure[\label{subfig:Afrikaans}Afrikaans]{\includegraphics[width=0.325\columnwidth]{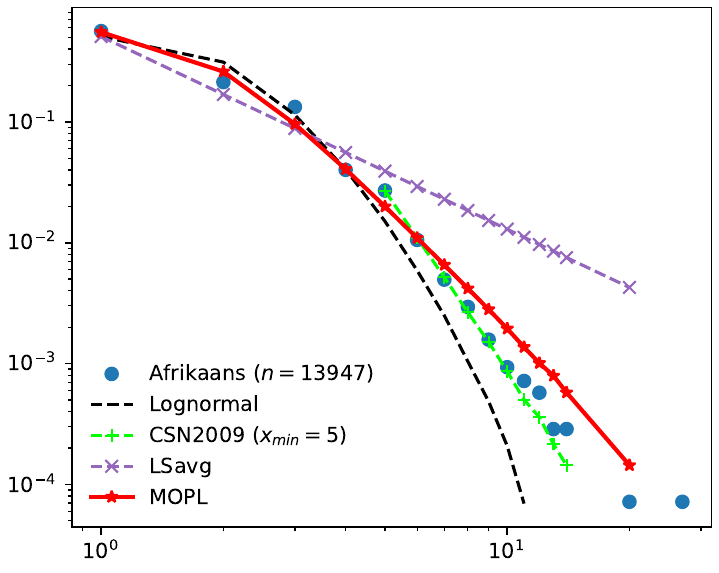}}
\subfigure[\label{subfig:Arabic}Arabic]{\includegraphics[width=0.325\columnwidth]{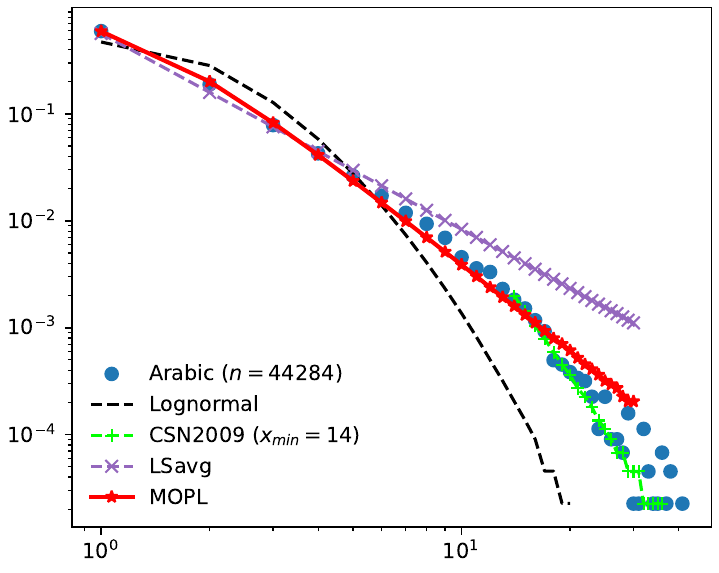}}
\subfigure[\label{subfig:Basque}Basque]{\includegraphics[width=0.325\columnwidth]{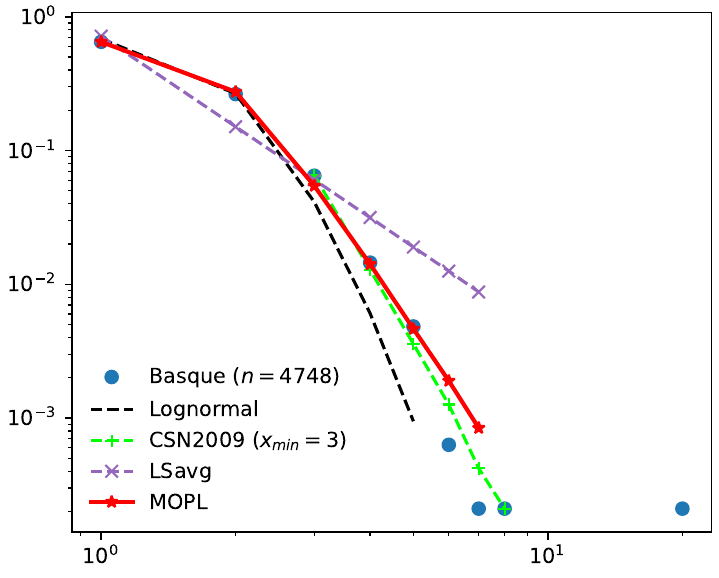}}
\subfigure[\label{subfig:Bokmal}Bokmal]{\includegraphics[width=0.325\columnwidth]{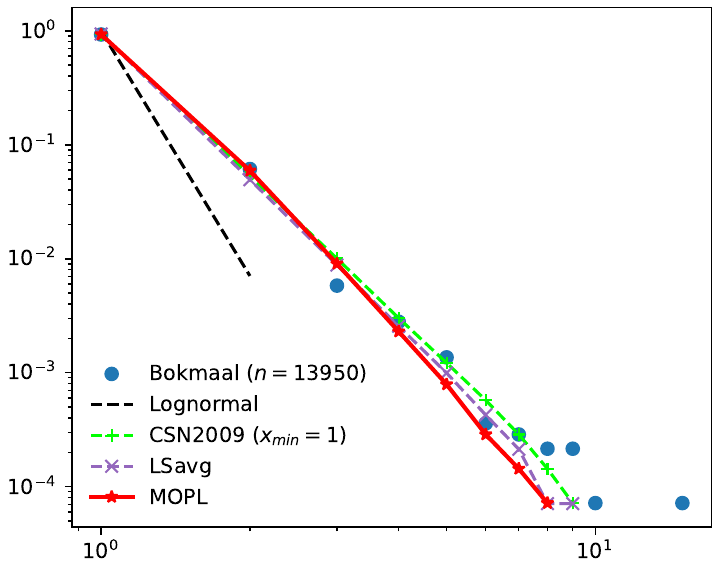}}
\subfigure[\label{subfig:Croatian}Croatian]{\includegraphics[width=0.325\columnwidth]{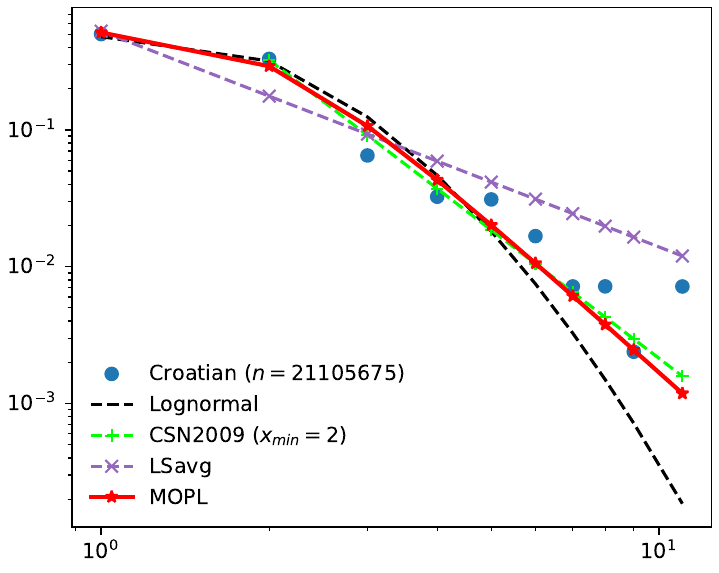}}
\subfigure[\label{subfig:Czech}Czech]{\includegraphics[width=0.325\columnwidth]{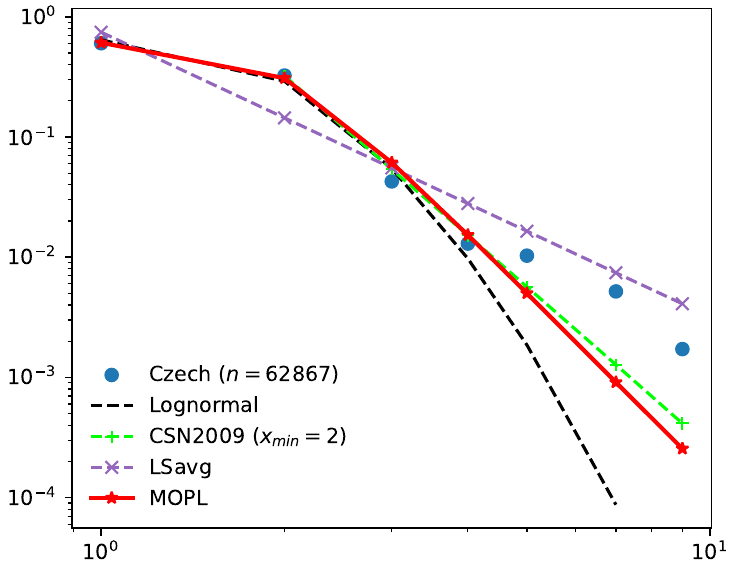}}
\subfigure[\label{subfig:France}France]{\includegraphics[width=0.325\columnwidth]{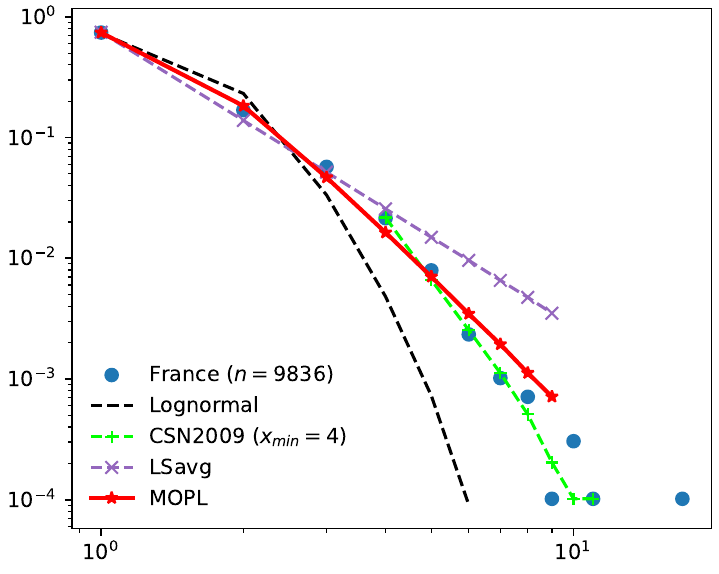}}
\subfigure[\label{subfig:German}German]{\includegraphics[width=0.325\columnwidth]{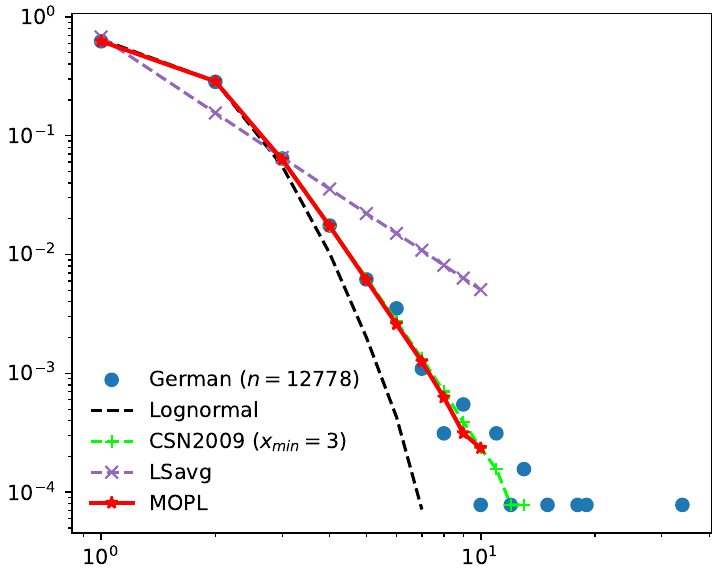}}
\subfigure[\label{subfig:Italian}Italian]{\includegraphics[width=0.325\columnwidth]{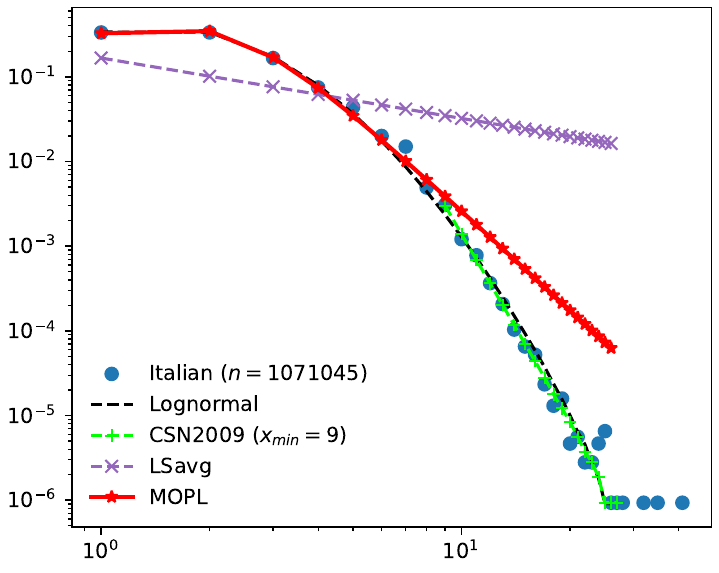}}
\caption{Plots of MOPL and the three compared models fitting to the length-frequency distributions of entities in different languages in the first nine datasets. The horizontal axis indicates the entity length ($l$) while the vertical axis indicates the percentage ($p(l)$).}
\label{fig:fitting-language-1}
\end{figure}

\begin{figure}[!htp]
\subfigure[\label{subfig:Netherland}Netherland]{\includegraphics[width=0.325\columnwidth]{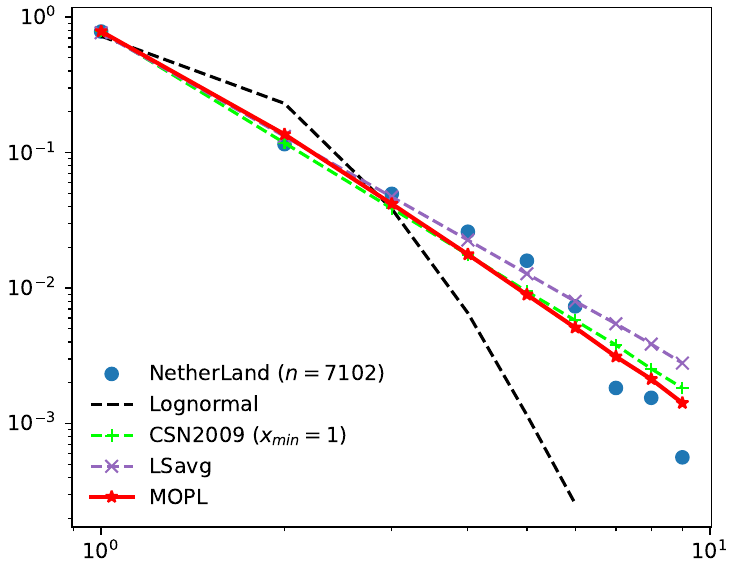}}
\subfigure[\label{subfig:Nynorsk}Nynorsk]{\includegraphics[width=0.325\columnwidth]{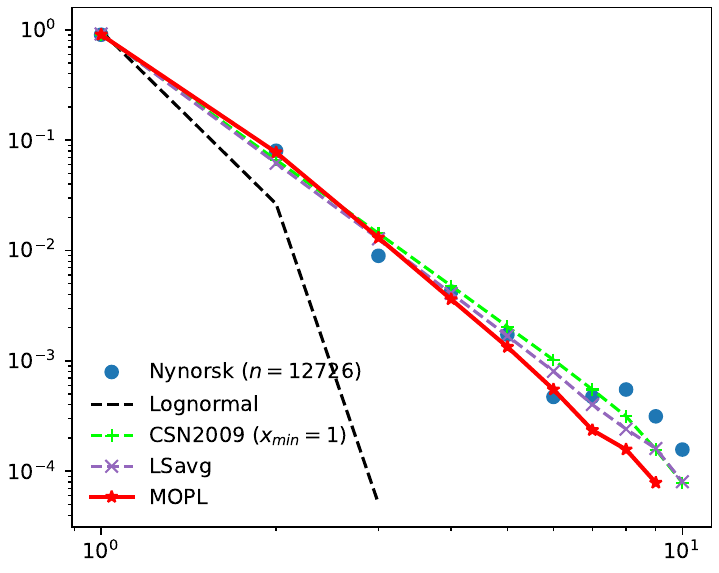}}
\subfigure[\label{subfig:Polish}Polish]{\includegraphics[width=0.325\columnwidth]{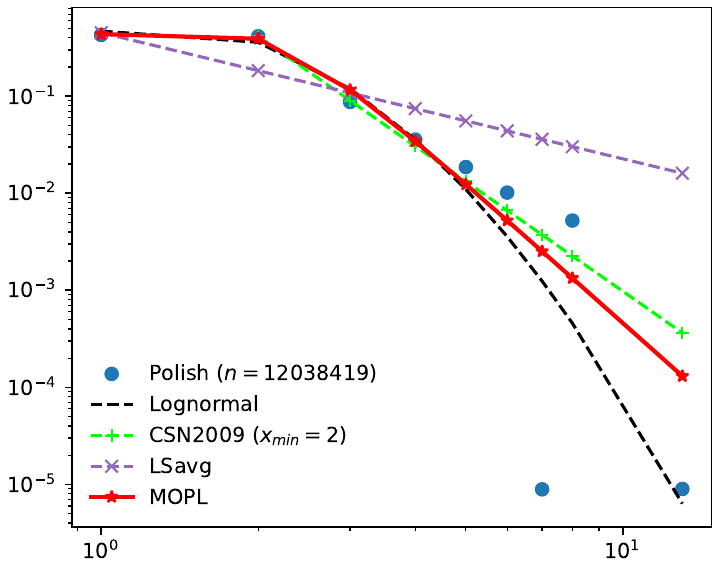}}
\subfigure[\label{subfig:Romanian}Romanian]{\includegraphics[width=0.325\columnwidth]{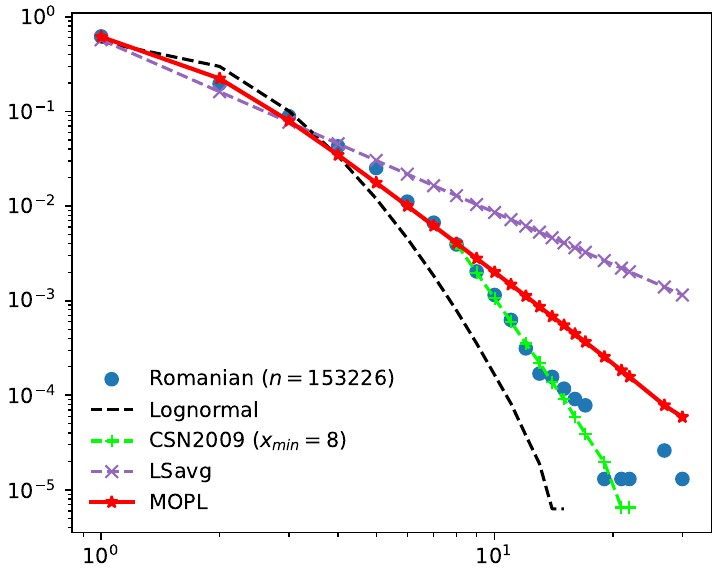}}
\subfigure[\label{subfig:Russian}Russian]{\includegraphics[width=0.325\columnwidth]{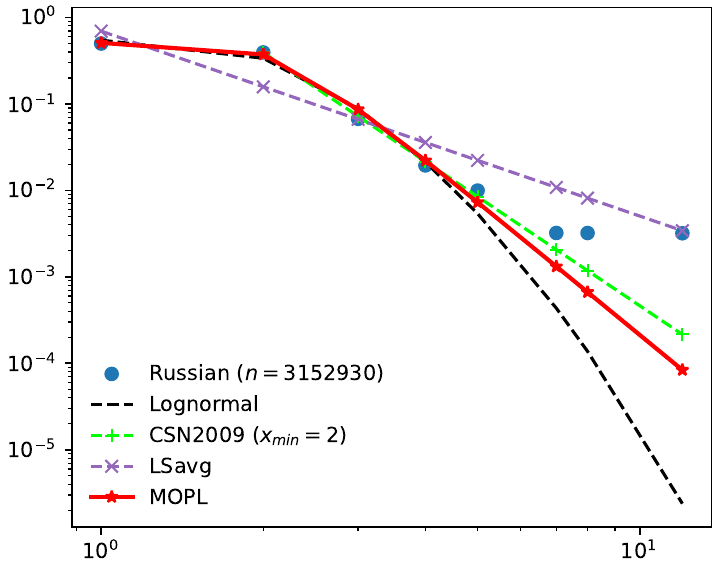}}
\subfigure[\label{subfig:Samnorsk}Samnorsk]{\includegraphics[width=0.325\columnwidth]{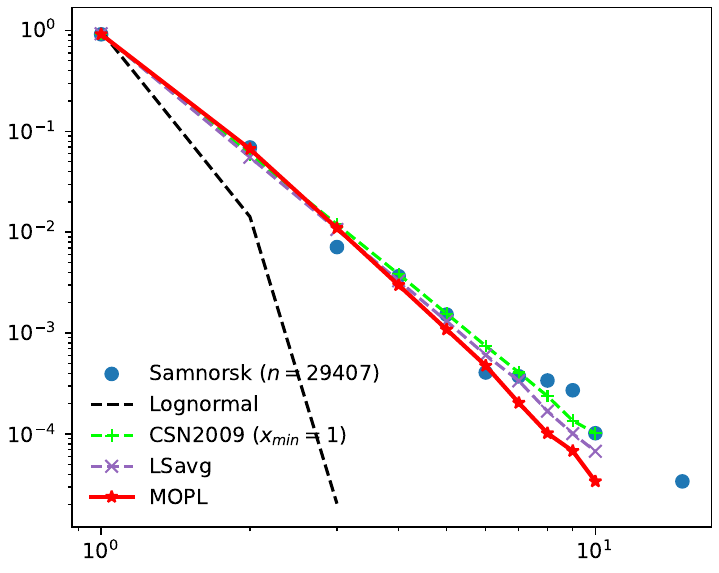}}
\subfigure[\label{subfig:Slovak}Slovak]{\includegraphics[width=0.325\columnwidth]{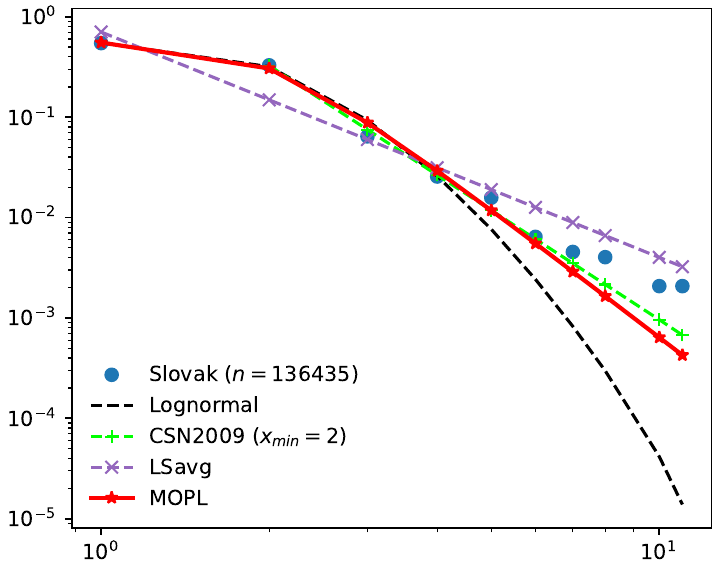}}
\subfigure[\label{subfig:Slovene}Slovene]{\includegraphics[width=0.325\columnwidth]{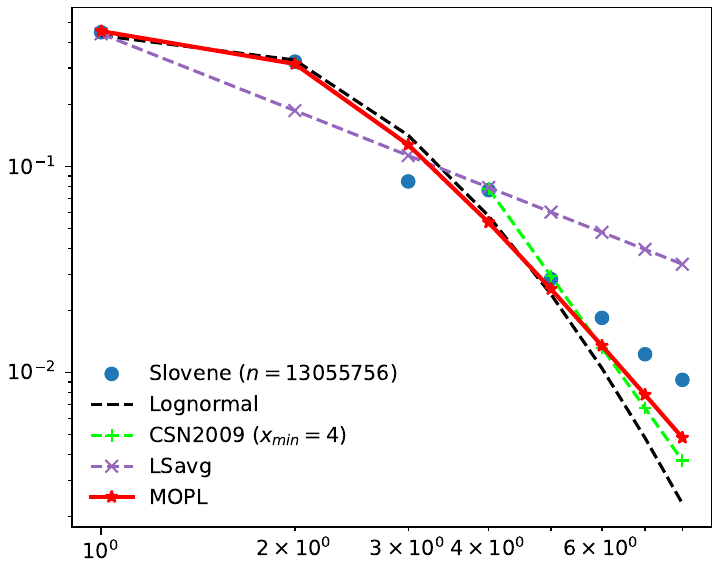}}
\subfigure[\label{subfig:Ukrainian}Ukrainian]{\includegraphics[width=0.325\columnwidth]{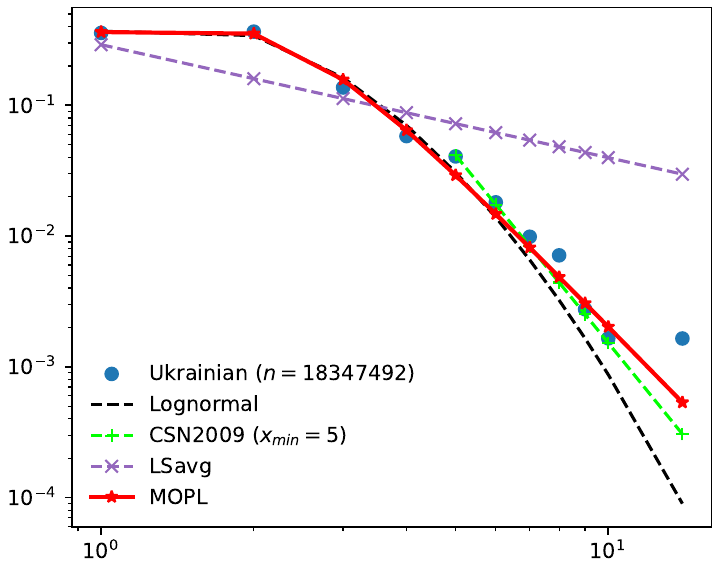}}
\caption{Plots of MOPL and the three compared models fitting to the length-frequency distributions of entities in different languages in the remaining nine datasets. The horizontal axis indicates the entity length ($l$) while the vertical axis indicates the percentage ($p(l)$).}
\label{fig:fitting-language-2}
\end{figure}

Tables~\ref{tb:parameter-different-type} and~\ref{tb:result-different-type} report the fitting and goodness-of-fit testing results of MOPL and the three compared models on the length-frequency distributions of entities in different types. Specifically, Table~\ref{tb:parameter-different-type} reports the estimated parameters of the models and the coverages (i.e., percentages of data that models cover) while Table~\ref{tb:result-different-type} reports the goodness-of-fit testing results of the models on the datasets, including $D_n$, $E_{avg}$, and $DEC$ where $DEC$ indicates the decision to accept or reject the hypothesis $H_0$. Figure~\ref{fig:fitting-type} visualizes the results of MOPL and the three compared models fitting to the length-frequency distributions of entities in different types. Tables~\ref{tb:parameter-different-language} reports the fitting results while Table~\ref{tb:result-different-language} reports the goodness-of-fit testing results of MOPL and the three compared models fitting to the length-frequency of entities in different languages. Figures~\ref{fig:fitting-language-1} and~\ref{fig:fitting-language-2} visualize those fittings to the length-frequency of entities in different languages.

What follows are separate discussions on model fitting and testing results on the length-frequency of entities in different types and different languages.

\subsubsection{Results on the length-frequency of entities in different types}\label{sssec:fitting-result-type}

Let us first look at the three measures that examine the goodness-of-fit in Table~\ref{tb:result-different-type}: $D_n$, $E_{avg}$, and $DEC$. Table~\ref{tb:result-different-type} shows that MOPL achieves the best results in all the three measures on all the twelve datasets, in comparison with the three compared models. Specifically, MOPL achieves the performance of $D_n$ in the range from 7.88E-05 to 1.22E-02 and the $E_{avg}$ value from 0.18 to 1.40 as well as all the ``\textbf{Accept}'' across the twelve datasets. By contrast, $LS_{avg}$ achieves the performance of $D_n$ in the range from 2.73E-01 to 8.00E-01 and the $E_{avg}$ value from 1.12 to 4.57 as well as all the ``Reject'' across the datasets. The three measures that CSN2009 achieves are 4.46E-03$\sim$6.02E-02 for $D_n$, 0.25$\sim$0.66 for $E_{avg}$, and 5 ``\textbf{Accept}'' and 7 ``Reject'' for $DEC$. The three measures of LogNormal are 1.76E-02$\sim$1.21E-01 for $D_n$, 0.36$\sim$11.27 for $E_{avg}$, and all 12 ``Reject'' for $DEC$. This indicates that MOPL fits the length-frequency distributions of entities in different types much better than $LS_{avg}$ and CSN2009, which are developed to fit power-law distributions, and LogNormal, which is often used as an alternative model for power-law models to fit empirical data. Figure~\ref{fig:fitting-type} intuitively visualizes the difference between MOPL and the three compared models in fitting the length-frequency distributions of entities on the twelve datasets. From Figure~\ref{fig:fitting-type} we can see that the fittings of MOPL are much better than the ones of the three compared models. More importantly, MOPL achieving all the ``\textbf{Accept}'' on the twelve datasets indicates that MOPL is a suitable model to characterize the length-frequency of entities in different types.

The fact that MOPL achieves the best goodness-of-fit testing results indicates that MOPL achieves the best estimated parameters. As shown in Table~\ref{tb:parameter-different-type}, therefore, the $\hat\alpha$ of MOPL should be considered as the relatively accurate estimated exponents fitting to the power-law segments of the length-frequency distributions of entities in different types. All the $\hat\alpha$ of MOPL fitting to these different types of entities range from 2.69 to 5.83, and most of these $\hat\alpha$ range from 2.69 to 4.74. This indicates that the length-frequency of entities in different types have stable scaling property.

\begin{table}[t]
\small
\centering
\caption{Results of MOPL and compared models fitting to the length-frequency distributions of entities in different languages. $C$ indicates the coverage which is defined by the percentage of data covered by a model. $M_{log}$ denotes logarithmic mean while $V_{log}$ denotes logarithmic variance.}
\label{tb:parameter-different-language}
\begin{tabular}{
@{}p{1.45cm}
c@{\hspace{2.2mm}}r@{\hspace{2.2mm}}r
c@{\hspace{2.2mm}}r
c@{\hspace{1.8mm}}c@{\hspace{1mm}}r
c@{\hspace{2mm}}c@{\hspace{2mm}}r@{}}
\toprule
\multirow{2}{*}{\textbf{Dataset}} & \multicolumn{3}{c}{\textbf{MOPL}} & \multicolumn{2}{c}{$LS_{avg}$} & \multicolumn{3}{c}{\textbf{CSN2009}} & \multicolumn{3}{c}{\textbf{LogNormal}} \\
\cmidrule(lr){2-4}\cmidrule(lr){5-6}\cmidrule(lr){7-9}\cmidrule(lr){10-12}
    & $\hat\alpha$ & $\hat\beta$	&	$C$(\%) & $\hat\alpha$	&	$C$(\%)  & $\hat{\alpha}$	& $\hat{x}_{min}$	&	$C$(\%)  & $M_{log}$ & $V_{log}$	&	$C$(\%)  \\
\midrule
Afrikaans	&	3.42&6.01&99.63	&	1.59&99.99	&	4.90&5&4.92	&0.44&0.31&100.00\\
Arabic	&	2.66&3.02&99.57	&	2.25&99.96	&	4.72&14&0.80	&0.47&0.45&100.00\\
Basque	&	4.91&13.74&99.77	&	4.25&99.96	&	5.60&3&8.34	&0.29&0.17&100.00\\
Bokmal	&	4.69&1.66&99.71	&	1.58&99.99	&	4.12&1&99.71	&0.06&0.05&100.00\\
Croatian	&	3.67&8.78&99.40	&	2.37&100.00	&	3.12&2&49.58	&0.48&0.32&100.00\\
Czech	&	5.08&18.68&99.70	&	1.98&100.00	&	4.41&2&39.92	&0.32&0.18&100.00\\
France	&	3.83&3.73&99.69	&	2.12&99.95	&	5.30&4&3.29	&0.23&0.18&100.00\\
German	&	4.74&13.38&99.82	&	1.09&99.91	&	4.53&3&9.38	&0.31&0.19&100.00\\
Italian	&	3.91&23.10&99.95	&	0.71&100.00	&	7.35&9&0.60	&0.68&0.33&100.00\\
Netherland&	3.06&1.49&99.34	&	3.89&100.00	&	2.74&1&98.47	&0.22&0.20&100.00\\
Nynorsk	&	4.49&1.95&99.94	&	1.30&100.00	&	3.77&1&88.37	&0.08&0.06&100.00\\
Polish	&	4.79&29.87&99.79	&	1.82&100.00	&	3.76&2&56.15	&0.49&0.23&100.00\\
Romanian	&	3.21&3.81&99.80	&	2.14&100.00	&	5.94&8&0.85	&0.39&0.30&100.00\\
Russian	&	5.12&28.91&99.62	&	4.06&100.00	&	4.19&2&49.85	&0.41&0.21&100.00\\
Samnorsk	&	4.53&1.70&99.98	&	2.25&100.00	&	3.95&1&99.63	&0.07&0.05&100.00\\
Slovak	&	4.24&12.01&99.77	&	1.24&100.00	&	3.62&2&45.30	&0.40&0.25&100.00\\
Slovene	&	3.68&11.37&98.77	&	0.86&100.00	&	4.38&4&13.11	&0.54&0.33&100.00\\
Ukrainian	&	3.98&21.16&99.47	&	1.83&100.00	&	4.77&5&7.60	&0.63&0.32&100.00\\
\bottomrule
\end{tabular}
\end{table}

\begin{sidewaystable}[htp]
\small
\centering
\caption{Goodness-of-fit testing results of MOPL and compared models fitting to the length-frequency distributions of entities in different languages. $D_{n}$ indicates the KS statistic defined by Eq.~(\ref{eq:kstest}). $E_{avg}$ indicates the average error defined by Eq.~(\ref{eq:avgerr}). $DEC$ indicates the decision to accept or reject the hypothesis $H_0$ that a model well fits the data, based on the $p$-value of the KS test. For each of $D_{n}$ and $E_{avg}$, the best result on each dataset is highlighted in bold.}
\label{tb:result-different-language}
\begin{tabular}{
  @{}l
  p{1.6cm}@{\hspace{2mm}}
  p{0.6cm}
  p{1.1cm}
  p{1.6cm}@{\hspace{2mm}}
  p{0.6cm}
  p{1.1cm}
  p{1.6cm}@{\hspace{2mm}}
  p{0.6cm}
  p{1.1cm}
  p{1.6cm}@{\hspace{2mm}}
  p{0.6cm}
  p{1.1cm}
  @{}
}
\toprule
\multirow{2}{*}{\textbf{Dataset}}	&	\multicolumn{3}{c}{\textbf{MOPL}}	&	\multicolumn{3}{c}{$LS_{avg}$}&\multicolumn{3}{c}{\textbf{CSN2009}}&	\multicolumn{3}{c}{\textbf{LogNormal}}		\\
\cmidrule(lr){2-4}\cmidrule(lr){5-7}\cmidrule(lr){8-10}\cmidrule(lr){11-13}
	&	\centering{$D_n$}	&	$E_{avg}$	&	$DEC$	&	\centering{$D_n$}	&	$E_{avg}$	&	$DEC$	
	&	\centering{$D_n$}	&	$E_{avg}$	&	$DEC$	&	\centering{$D_n$}	&	$E_{avg}$	&	$DEC$	\\
\midrule
Afrikaans&\textbf{1.72E-03}&0.42&\textbf{Accept}&4.67E-01&2.16&Reject&2.24E-02&\textbf{0.22}&\textbf{Accept}&6.53E-02&0.86&Reject\\
Arabic&\textbf{6.07E-03}&\textbf{0.37}&\textbf{Accept}&4.33E-01&1.41&Reject&5.66E-02&0.39&\textbf{Accept}&1.24E-01&1.80&Reject\\
Basque&1.50E-02&\textbf{0.24}&\textbf{Accept}&2.86E-01&1.21&Reject&\textbf{7.06E-03}&0.31&\textbf{Accept}&8.63E-02&0.65&Reject\\
Bokmal&\textbf{1.34E-02}&0.41&Reject&2.00E-01&0.43&Reject&5.41E-02&\textbf{0.32}&Reject&4.69E-02&1.34&Reject\\
Croatian&\textbf{1.53E-02}&0.30&Reject&3.00E-01&0.80&Reject&2.08E-02&\textbf{0.29}&Reject&5.88E-02&0.70&Reject\\
Czech&\textbf{4.01E-02}&0.55&Reject&1.43E-01&\textbf{0.49}&Reject&5.69E-02&1.89&Reject&4.60E-02&1.70&Reject\\
France&\textbf{2.13E-03}&\textbf{0.27}&\textbf{Accept}&3.33E-01&0.87&Reject&4.92E-03&0.51&\textbf{Accept}&4.49E-02&1.73&Reject\\
German&\textbf{2.42E-03}&\textbf{0.20}&\textbf{Accept}&4.00E-01&1.69&Reject&2.18E-02&0.32&\textbf{Accept}&6.73E-02&1.16&Reject\\
Italian&\textbf{1.16E-02}&2.18&Reject&7.69E-01&23.99&Reject&3.47E-02&0.38&Reject&6.89E-02&\textbf{0.34}&Reject\\
Netherland&\textbf{8.98E-03}&0.32&\textbf{Accept}&2.22E-01&0.34&Reject&1.67E-02&\textbf{0.29}&Reject&7.06E-02&1.86&Reject\\
Nynorsk&\textbf{8.90E-03}&0.50&\textbf{Accept}&2.00E-01&\textbf{0.33}&Reject&2.17E-02&0.34&Reject&4.03E-02&4.81&Reject\\
Polish&2.04E-02&2.47&Reject&3.33E-01&8.78&Reject&5.21E-03&\textbf{0.35}&Reject&4.00E-02&2.12&Reject\\
Romanian&2.74E-02&\textbf{1.18}&Reject&5.45E-01&4.31&Reject&7.06E-03&3.18&\textbf{Accept}&3.72E-02&1.77&Reject\\
Russian&\textbf{5.51E-03}&0.49&Reject&1.25E-01&0.71&Reject&1.77E-02&\textbf{0.30}&Reject&4.03E-02&1.17&Reject\\
Samnorsk&\textbf{2.08E-03}&0.57&\textbf{Accept}&1.82E-01&0.36&Reject&1.52E-02&\textbf{0.25}&Reject&2.47E-02&6.81&Reject\\
Slovak&\textbf{9.13E-03}&0.40&Reject&1.00E-01&0.45&Reject&2.49E-02&\textbf{0.28}&Reject&5.55E-02&1.60&Reject\\
Slovene&3.63E-02&\textbf{0.24}&Reject&3.75E-01&0.56&Reject&8.79E-03&0.25&Reject&1.70E-02&0.37&Reject\\
Ukrainian&\textbf{2.26E-02}&0.17&Reject&4.55E-01&1.61&Reject&3.06E-02&\textbf{0.15}&Reject&7.39E-02&0.34&Reject\\
\bottomrule
\end{tabular}
\end{sidewaystable}

Let us now look at the fittings of the two state-of-the-art compared models, $LS_{avg}$ and CSN2009. The $\hat\alpha$ of $LS_{avg}$ are deviated relatively far away from the $\hat\alpha$ of MOPL. The reason is that $LS_{avg}$ assumes that a power-law starts from the very beginning of an empirical dataset, but Figure~\ref{fig:fitting-type} shows that such assumption is not applicable to the length-frequency of entities. This indicates that a pure power-law model is unsuitable to characterize the length-frequency of entities in different types. On the other hand, the $\hat\alpha$ of CSN2009 are deviated slightly from the the $\hat\alpha$ of MOPL. The reason is that CSN2009 adopts a minimum-KS-statistic strategy to choose larger lower bound (i.e., $\hat{x}_{min}$) and fits only the long tails. Consequently, CSN2009 discards the majority of data and achieves low coverages, which are only from 1.23\% to 70.99\%. By contrast, other models cover more than 98.70\% of data. This result that CSN2009 achieves low coverage in fitting to empirical data is consistent with the observation reported in~\citet{zhong2022least}.

\subsubsection{Results on the length-frequency of entities in different languages}\label{sssec:fitting-result-language}

Let us first look at the three goodness-of-fit testing measures in Table~\ref{tb:result-different-language} as well: $D_n$, $E_{avg}$, and $DEC$. Table~\ref{tb:result-different-language} shows that none of the four models (i.e., MOPL, $LS_{avg}$, CSN2009, and LogNormal) can perfectly characterize the length-frequency distributions of entities in the eighteen languages. The fittings to the length-frequency of entities in different languages are much worse than the fittings to the length-frequency of entities in different types. A possible reason is that some of these datasets in the non-English languages contain a large number of noises. As we mentioned above, English is the most studied language in the field of natural language processing and related areas; other languages are also studied, but their annotated datasets may not be as accurate as the datasets in English. Another possible reason is that none of our authors are familiar with those languages and cannot guarantee the accuracy of the annotations for these datasets.

Let us now look at the comparison among the four models fitting to the length-frequency of entities. While MOPL does not well characterize the length-frequency distributions of entities in all the eighteen languages, MOPL outperforms the three compared models. Specifically, MOPL achieves the $D_n$ value in the range from 1.72E-03 to 4.01E-02, achieves the $E_{avg}$ value in the range from 0.17 to 2.47, and achieves 8 ``\textbf{Accept}'' and 10 ``Reject'' for $DEC$ across all the eighteen languages. By contrast, $LS_{avg}$ achieves the $D_n$ value from 1.00E-01 to 7.69E-01, achieves the $E_{avg}$ value from 0.33 to 23.99, and achieves all 18 ``Reject'' for $DEC$ across the eighteen languages. CSN2009 achieves the $D_n$ value from 4.92E-03 to 5.69E-02, achieves the $E_{avg}$ value from 0.15 to 3.18, and achieves 6 ``\textbf{Accept}'' and 12 ``Reject'' for $DEC$. LogNormal achieves the $D_n$ value from 1.70E-02 to 1.24E-01, achieves the $E_{avg}$ value from 0.34 to 6.81,  and achieves all 18 ``Reject'' for $DEC$. The comparison among the four models fitting to the length-frequency of entities is intuitively visualized in Figures~\ref{fig:fitting-language-1} and~\ref{fig:fitting-language-2}. The fitting and testing results indicate that MOPL is more suitable to characterize the length-frequency distributions of entities in different languages than $LS_{avg}$, CSN2009, and LogNormal.

Table~\ref{tb:parameter-different-language} shows that the $\hat\alpha$ of MOPL fitting to the length-frequency distributions of entities in different languages range only from 2.66 to 5.12, which is consistent with the $\hat\alpha$ of MOPL fitting to different types of entities, as shown in Table~\ref{tb:parameter-different-type}. This indicates that the length-frequency distributions of entities in different languages also have stable scaling property. In terms of data coverage, MOPL, $LS_{avg}$, and LogNormal cover almost all the data (i.e., from 99.91\% to 100\%), while CSN2009 achieves relatively low coverages (i.e., lower to 0.60\%). Specifically, CSN2009 discards at least 50\% of data in 13 out of 18 languages, and discards at least 90\% of data in 8 out of 18 languages. The low coverage of CSN2009 on the length-frequency of entities in different languages is consistent with the one of CSN2009 on the length-frequency of entities in different types reported in Table~\ref{tb:parameter-different-type} as well as the observation reported in~\citet{zhong2022least}. 

\subsection{Computational Efficiency}\label{ssec:efficiency}

\begin{table}[htp]
\centering
\caption{Runtime of MOPL, $LS_{avg}$, CSN2009, and LogNormal fitting to the length-frequency distributions of entities in different types and different languages. The unit of the runtime is millisecond, denoted by $ms$.}
\label{tb:runtime}
\begin{tabular}{@{}lrrrr@{}}
\toprule
\textbf{Dataset}	&	\textbf{MOPL}	&	\textbf{$LS_{avg}$}	&	\textbf{CSN2009}	&	\textbf{LogNormal}	\\
\midrule
ABSA		&	188.93 $ms$	&	5.89 $ms$		&	29.51 $ms$	&	6.20 $ms$	\\
ACE04		&	293.97 $ms$	&	6.40 $ms$		&	308.19 $ms$	&	7.14 $ms$	\\
BBN			&	69.83 $ms$	&	6.81 $ms$		&	134.39 $ms$	&	6.32 $ms$	\\
BioMed		&	360.48 $ms$	&	7.03 $ms$		&	4368.31 $ms$	&	7.43 $ms$	\\
CoNLL03		&	360.48 $ms$	&	5.71 $ms$		&	42.93 $ms$	&	6.92 $ms$	\\
COVID19		&	261.38 $ms$	&	7.52 $ms$		&	39544.32 $ms$	&	27.45 $ms$	\\
LitBank		&	409.67 $ms$	&	6.78 $ms$		&	474.60 $ms$	&	6.57 $ms$	\\
OntoNotes5	&	96.58 $ms$	&	5.60 $ms$		&	183.25 $ms$	&	8.53 $ms$	\\
Re3d			&	111.97 $ms$	&	6.20 $ms$		&	19.79 $ms$	&	6.90 $ms$	\\
TimeExp		&	137.48 $ms$	&	6.54 $ms$		&	59.12 $ms$	&	6.66 $ms$	\\
Twitter		&	89.37 $ms$	&	152.74 $ms$	&	53.19 $ms$	&	1371.74 $ms$	\\
WikiAnchor	&	357.21 $ms$	&	7.05 $ms$		&	17060.66 $ms$	&	12.55 $ms$\\
\textbf{Total}	&\textbf{2737.35} $ms$	&\textbf{224.27} $ms$	&\textbf{62278.26} $ms$	&\textbf{1474.41}$ms$	\\
\midrule
Afrikaans	&	312.27 $ms$	&	6.34 $ms$	&	53.83	 $ms$	&	6.58 $ms$	\\
Arabic	&	224.97 $ms$	&	7.13 $ms$	&	284.04 $ms$	&	6.68 $ms$	\\
Basque	&	64.78 $ms$	&	6.44 $ms$	&	13.29	 $ms$	&	6.30 $ms$	\\
Bokmaal	&	92.05 $ms$	&	6.13 $ms$	&	22.85 $ms$	&	6.03 $ms$	\\
Croatian	&	73.45 $ms$	&	6.09 $ms$	&	31483.92 $ms$	&	88.09 $ms$	\\
Czech	&	69.13 $ms$	&	6.50 $ms$	&	80.67	 $ms$	&	6.09 $ms$	\\
France	&	79.26 $ms$	&	6.48 $ms$	&	23.68 $ms$	&	7.02 $ms$	\\
German	&	168.32 $ms$	&	227.47 $ms$&	88.78 $ms$	&	783.02 $ms$	\\
Italian	&	295.43 $ms$	&	6.26 $ms$	&	6335.01 $ms$	&	9.42 $ms$	\\
Netherland	&	41.71 $ms$	&	6.84 $ms$	&	11.21 $ms$	&	6.37 $ms$	\\
Nynorsk	&	69.92	 $ms$	&	6.28 $ms$	&	21.86 $ms$	&	6.61 $ms$	\\
Polish	&	67.35	 $ms$	&	5.47 $ms$	&	20347.38 $ms$	&	99.88 $ms$	\\
Romanian	&	132.39 $ms$	&	6.20 $ms$	&	527.88 $ms$	&	6.26 $ms$	\\
Russian	&	82.65 $ms$	&	6.06 $ms$	&	4555.56 $ms$	&	12.21 $ms$	\\
Samnorsk	&	89.67	 $ms$	&	5.80 $ms$	&	41.98 $ms$	&	6.03 $ms$	\\
Slovak	&	114.66 $ms$	&	6.12 $ms$	&	185.98 $ms$	&	6.17 $ms$	\\
Slovene	&	60.35	 $ms$	&	6.30 $ms$	&	15422.35 $ms$	&	39.23 $ms$	\\
Ukrainian	&	94.12 $ms$	&	7.39 $ms$	&	37443.65 $ms$	&	50.21 $ms$	\\
\textbf{Total}	&\textbf{2132.46} $ms$	&	\textbf{335.30} $ms$	&	\textbf{116943.92} $ms$	&	\textbf{1152.21} $ms$	\\

\bottomrule
\end{tabular}
\end{table}

Table~\ref{tb:runtime} reports the runtimes of MOPL, $LS_{avg}$, CSN2009 and LogNormal fitting to the length-frequency distributions of entities in different types and different languages.\ftext{Note that the reported runtimes only include the time of the four models fitting to the length-frequency distributions; they do not include the time of the KS testing.} Table~\ref{tb:runtime} shows that while the runtimes of MOPL fitting to length-frequency of entities in both different types and different languages are less efficient than ones of $LS_{avg}$ and LogNormal, they are significantly more efficient than the ones of CSN2009. Moreover, while the number of entities in individual dataset ranges from 3,394 to 10,260,797 in different types (see Table~\ref{tb:dataset-different-type}) and from 4,748 to 21,105,675 in different languages (see Table~\ref{tb:dataset-different-language}), the runtime of MOPL performing on individual dataset ranges only from 41.71 to 409.67 milliseconds, all of which are less than one second. That means the runtime of MOPL neither increases linearly nor exponentially as the number of entities increases. This suggests that MOPL can be easily applied on large-scale datasets with high efficiency.

\section{Discussion}\label{sec:discussion}

\subsection{Some Implications on Entity-related Linguistic Tasks}\label{ssec:implications}

We here briefly discuss some implications of this linguistic phenomenon (i.e., the length-frequency of entities in different types and different languages can be characterized by Marshall-Olkin power-law distributions) on entity-related linguistic tasks. This linguistic phenomenon may be able to explain why many statistical models and deep-learning models, such as conditional random fields~\citep{CRF2001}, long short-term memory networks~\citep{LSTM1997}, and transformer~\citep{devlin2018bert}, can be applied for recognizing all these different types of entities from unstructured text~\citep{FukudaEtal1998, CoNLL2003, TakeuchiAndCollier2005, NERSurvey2007, RitterEtal2011, Liu2012, PontikiEtal2014, KrallingerEtal2013, BTC16, NER2018Survey, MyThesis2020, ZhongEtal2022syntax}. This linguistic phenomenon may also be able to provide insights into analyzing those languages with low-resources. Since entities in different types and different languages share many common characteristics (e.g., their length-frequency distributions, average lengths, and scaling property), we could transfer knowledge and resource available in those well-studied languages to those low-resource languages. We could also apply those statistical modes and deep-learning models that have demonstrated to be effective and efficient in well-studied languages to those low-resource languages. Distilling this knowledge about the length-frequency distributions of entities can also drive us to design effective and efficient algorithms for specific linguistic tasks. For example, \citet{SynTime2017} found that an average time expression contains only about two words of which one is time token and the other is modifier or numeral, and then they designed proper rules to recognize time expressions from unstructured text. To apply this linguistic knowledge and achieve more progress in linguistic tasks, however, we still need to explore into deeper understanding of this linguistic phenomenon.

\subsection{Limitations}\label{ssec:limitations}

While we find that the length-frequency distributions of entities in different types can be well characterized by Marshall-Olkin power-law (MOPL) models, and the ones in different languages can also be roughly characterized by MOPL models, we should note that our analysis on these datasets about different languages may be inaccurate because many of these languages are not well studied in the field of natural language processing and related areas and we authors do not have sufficient expertise knowledge to cover our analysis on these different languages.

\section{Conclusion}\label{sec:conclusion}

In this paper, we discover that the length-frequency distributions of entities in different types and different languages can be characterized by a family of Marshall-Olkin power-law (MOPL) models. Our discovery adds a stable knowledge to the field of language and provides some insights into conducting entity-related linguistic tasks and may also provide a new perspective for future potential research in understanding the language use. Experimental results on the length-frequency of entities in both different types and different languages demonstrate the superiority of MOPL models against a log-normal model and two state-of-the-art power-law models, namely $LS_{avg}$ that is developed by~\citet{zhong2022least} and CSN2009 that is developed by~\citet{ClausetEtal2009}. Experimental results also demonstrate that MOPL models are scalable to the length-frequency of entities in large-scale real-world datasets.

\section*{Acknowledgments}
This research is supported by the Agency for Science, Technology and Research (A*STAR) under its AME Programmatic Funding Scheme (Project \#A18A2b0046).


\bibliographystyle{elsarticle-harv}
\bibliography{powerlaw}

\begin{thebibliography}{78}
\expandafter\ifx\csname natexlab\endcsname\relax\def\natexlab#1{#1}\fi
\providecommand{\url}[1]{\texttt{#1}}
\providecommand{\href}[2]{#2}
\providecommand{\path}[1]{#1}
\providecommand{\DOIprefix}{doi:}
\providecommand{\ArXivprefix}{arXiv:}
\providecommand{\URLprefix}{URL: }
\providecommand{\Pubmedprefix}{pmid:}
\providecommand{\doi}[1]{\href{http://dx.doi.org/#1}{\path{#1}}}
\providecommand{\Pubmed}[1]{\href{pmid:#1}{\path{#1}}}
\providecommand{\bibinfo}[2]{#2}
\ifx\xfnm\relax \def\xfnm[#1]{\unskip,\space#1}\fi
\bibitem[{Anbalagan et~al.(2021)Anbalagan, Hincal, Ramachandran, Baleanu, Cao
  and Niezabitowski}]{anbalagan2021razumikhin}
\bibinfo{author}{Anbalagan, P.}, \bibinfo{author}{Hincal, E.},
  \bibinfo{author}{Ramachandran, R.}, \bibinfo{author}{Baleanu, D.},
  \bibinfo{author}{Cao, J.}, \bibinfo{author}{Niezabitowski, M.},
  \bibinfo{year}{2021}.
\newblock \bibinfo{title}{A razumikhin approach to stability and
  synchronization criteria for fractional order time delayed gene regulatory
  networks} \bibinfo{volume}{6}, \bibinfo{pages}{4526--4555}.
\bibitem[{Arnold and Emerson(2011)}]{arnold2011nonparametric}
\bibinfo{author}{Arnold, T.B.}, \bibinfo{author}{Emerson, J.W.},
  \bibinfo{year}{2011}.
\newblock \bibinfo{title}{Nonparametric goodness-of-fit tests for discrete null
  distributions.}
\newblock \bibinfo{journal}{R Journal} \bibinfo{volume}{3}.
\bibitem[{Artico et~al.(2020)Artico, Smolyarenko, Vinciotti and
  Wit}]{ArticoEtal2020}
\bibinfo{author}{Artico, I.}, \bibinfo{author}{Smolyarenko, I.},
  \bibinfo{author}{Vinciotti, V.}, \bibinfo{author}{Wit, E.C.},
  \bibinfo{year}{2020}.
\newblock \bibinfo{title}{How rare are power-law networks really?}, in:
  \bibinfo{booktitle}{Proceedings of the Royal Society A}, p.
  \bibinfo{pages}{20190742}.
\bibitem[{Bamman et~al.(2019)Bamman, Popat and Shen}]{Litbank2019}
\bibinfo{author}{Bamman, D.}, \bibinfo{author}{Popat, S.},
  \bibinfo{author}{Shen, S.}, \bibinfo{year}{2019}.
\newblock \bibinfo{title}{An annotated dataset of literary entities}, in:
  \bibinfo{booktitle}{Proceedings of the 2019 Conference of the North American
  Chapter of the Association for Computational Linguistics: Human Language
  Technologies}, pp. \bibinfo{pages}{2138--2144}.
\bibitem[{Best(1996)}]{Best1996}
\bibinfo{author}{Best, K.H.}, \bibinfo{year}{1996}.
\newblock \bibinfo{title}{Word length in old icelandic songs and prose texts}.
\newblock \bibinfo{journal}{Journal of Quantitative Linguistics}
  \bibinfo{volume}{3}, \bibinfo{pages}{97--105}.
\bibitem[{Chinchor(1997)}]{MUC7}
\bibinfo{author}{Chinchor, N.A.}, \bibinfo{year}{1997}.
\newblock \bibinfo{title}{Muc-7 named entity task definition}, in:
  \bibinfo{booktitle}{Proceedings of the 7th Message Understanding Conference}.
\bibitem[{Clauset et~al.(2009)Clauset, Shalizi and Newman}]{ClausetEtal2009}
\bibinfo{author}{Clauset, A.}, \bibinfo{author}{Shalizi, C.R.},
  \bibinfo{author}{Newman, M.E.J.}, \bibinfo{year}{2009}.
\newblock \bibinfo{title}{Power-law distributions in empirical data}.
\newblock \bibinfo{journal}{SIAM Review} \bibinfo{volume}{51},
  \bibinfo{pages}{661--703}.
\bibitem[{Corominas-Murtra and Sol{\'e}(2010)}]{BernatRicard2010}
\bibinfo{author}{Corominas-Murtra, B.}, \bibinfo{author}{Sol{\'e}, R.V.},
  \bibinfo{year}{2010}.
\newblock \bibinfo{title}{Universality of zipf's law}.
\newblock \bibinfo{journal}{Physical Review E} \bibinfo{volume}{82},
  \bibinfo{pages}{011102}.
\bibitem[{Crichton et~al.(2017)Crichton, Pyysalo, Chiu and
  Korhonen}]{CrichtonEtal2017}
\bibinfo{author}{Crichton, G.}, \bibinfo{author}{Pyysalo, S.},
  \bibinfo{author}{Chiu, B.}, \bibinfo{author}{Korhonen, A.},
  \bibinfo{year}{2017}.
\newblock \bibinfo{title}{A neural network multi-task learning approach to
  biomedical named entity recognition}.
\newblock \bibinfo{journal}{BMC Bioinformatics} \bibinfo{volume}{18},
  \bibinfo{pages}{368--371}.
\bibitem[{Derczynski et~al.(2016)Derczynski, Bontcheva and Roberts}]{BTC16}
\bibinfo{author}{Derczynski, L.}, \bibinfo{author}{Bontcheva, K.},
  \bibinfo{author}{Roberts, I.}, \bibinfo{year}{2016}.
\newblock \bibinfo{title}{Broad twitter corpus: A diverse named entity
  recognition resource}, in: \bibinfo{booktitle}{Proceedings of the 26th
  International Conference on Computational Linguistics}, pp.
  \bibinfo{pages}{1169--1179}.
\bibitem[{Devlin et~al.(2018)Devlin, Chang, Lee and Toutanova}]{devlin2018bert}
\bibinfo{author}{Devlin, J.}, \bibinfo{author}{Chang, M.W.},
  \bibinfo{author}{Lee, K.}, \bibinfo{author}{Toutanova, K.},
  \bibinfo{year}{2018}.
\newblock \bibinfo{title}{Bert: Pre-training of deep bidirectional transformers
  for language understanding}.
\newblock \bibinfo{journal}{arXiv preprint arXiv:1810.04805} .
\bibitem[{Dimitrova et~al.(2020)Dimitrova, Kaishev and Tan}]{JSSv095i10}
\bibinfo{author}{Dimitrova, D.S.}, \bibinfo{author}{Kaishev, V.K.},
  \bibinfo{author}{Tan, S.}, \bibinfo{year}{2020}.
\newblock \bibinfo{title}{Computing the kolmogorov-smirnov distribution when
  the underlying cdf is purely discrete, mixed, or continuous}.
\newblock \bibinfo{journal}{Journal of Statistical Software}
  \bibinfo{volume}{95}, \bibinfo{pages}{1--42}.
\newblock \URLprefix
  \url{https://www.jstatsoft.org/index.php/jss/article/view/v095i10},
  \DOIprefix\doi{10.18637/jss.v095.i10}.
\bibitem[{Doddington et~al.(2004)Doddington, Mitchell, Przybocki, Ramshaw,
  Strassel and Weischedel}]{ACE2004}
\bibinfo{author}{Doddington, G.}, \bibinfo{author}{Mitchell, A.},
  \bibinfo{author}{Przybocki, M.}, \bibinfo{author}{Ramshaw, L.},
  \bibinfo{author}{Strassel, S.}, \bibinfo{author}{Weischedel, R.},
  \bibinfo{year}{2004}.
\newblock \bibinfo{title}{The automatic content extraction (ace) program tasks,
  data, and evaluation}, in: \bibinfo{booktitle}{Proceedings of the 2004
  Conference on Language Resources and Evaluation}, pp. \bibinfo{pages}{1--4}.
\bibitem[{Dumitrescu and Avram(2019)}]{dumitrescu2019introducing}
\bibinfo{author}{Dumitrescu, S.D.}, \bibinfo{author}{Avram, A.M.},
  \bibinfo{year}{2019}.
\newblock \bibinfo{title}{Introducing ronec--the romanian named entity corpus}.
\newblock \bibinfo{journal}{arXiv preprint arXiv:1909.01247} .
\bibitem[{Estoup(1916)}]{Estoup1916}
\bibinfo{author}{Estoup, J.B.}, \bibinfo{year}{1916}.
\newblock \bibinfo{title}{Gammes stenographiques}, in:
  \bibinfo{booktitle}{Institut Stenographique de France, Paris}.
\bibitem[{Fucks(1955)}]{Fucks1955}
\bibinfo{author}{Fucks, W.}, \bibinfo{year}{1955}.
\newblock \bibinfo{title}{Theorie der wortbildung}.
\newblock \bibinfo{journal}{Mathematisch-Physikalische Semesterberichte}
  \bibinfo{volume}{4}, \bibinfo{pages}{195--212}.
\bibitem[{Fucks(1956)}]{Fucks1956}
\bibinfo{author}{Fucks, W.}, \bibinfo{year}{1956}.
\newblock \bibinfo{title}{Die mathematischen gesetze der bildung von
  sprachelementen aus ihren bestandteilen}.
\newblock \bibinfo{journal}{Nachrichtentechnische Fachberichte}
  \bibinfo{volume}{3}, \bibinfo{pages}{7--21}.
\bibitem[{Fukuda et~al.(1998)Fukuda, Tsunoda, Tamura and
  Takagi}]{FukudaEtal1998}
\bibinfo{author}{Fukuda, K.}, \bibinfo{author}{Tsunoda, T.},
  \bibinfo{author}{Tamura, A.}, \bibinfo{author}{Takagi, T.},
  \bibinfo{year}{1998}.
\newblock \bibinfo{title}{Toward information extraction: Identifying protein
  names from biological papes}, in: \bibinfo{booktitle}{Proceedings of the
  Pacific Symposium on Biocomputing}, pp. \bibinfo{pages}{707--718}.
\bibitem[{Gerlach and Altmann(2019)}]{GerlachAltmann2019}
\bibinfo{author}{Gerlach, M.}, \bibinfo{author}{Altmann, E.G.},
  \bibinfo{year}{2019}.
\newblock \bibinfo{title}{Testing statistical laws in complex systems}.
\newblock \bibinfo{journal}{Physical Review Letters} \bibinfo{volume}{122},
  \bibinfo{pages}{168301}.
\bibitem[{Grishman and Sundheim(1996)}]{MUC6}
\bibinfo{author}{Grishman, R.}, \bibinfo{author}{Sundheim, B.},
  \bibinfo{year}{1996}.
\newblock \bibinfo{title}{Message understanding conference - 6: A brief
  history}, in: \bibinfo{booktitle}{Proceedings of the 16th International
  Conference on Computational Linguistics}.
\bibitem[{Grotjahn and Altmann(1993)}]{GrotjahnAltmann1993}
\bibinfo{author}{Grotjahn, R.}, \bibinfo{author}{Altmann, G.},
  \bibinfo{year}{1993}.
\newblock \bibinfo{title}{Modelling the distribution of word length: Some
  methodological problems}.
\newblock \bibinfo{journal}{Contributions to Quatitative Linguistics} ,
  \bibinfo{pages}{141--153}.
\bibitem[{Hanel et~al.(2017)Hanel, Corominas-Murtra, Liu and
  Thurner}]{HanelEtal2017}
\bibinfo{author}{Hanel, R.}, \bibinfo{author}{Corominas-Murtra, B.},
  \bibinfo{author}{Liu, B.}, \bibinfo{author}{Thurner, S.},
  \bibinfo{year}{2017}.
\newblock \bibinfo{title}{Fitting power-laws in empirical data with estimators
  that work for all exponents}.
\newblock \bibinfo{journal}{PLoS ONE} \bibinfo{volume}{12},
  \bibinfo{pages}{1--15}.
\bibitem[{Hochreiter and Schmidhuber(1997)}]{LSTM1997}
\bibinfo{author}{Hochreiter, S.}, \bibinfo{author}{Schmidhuber, J.},
  \bibinfo{year}{1997}.
\newblock \bibinfo{title}{Long short-term memory}.
\newblock \bibinfo{journal}{Neural Computation} \bibinfo{volume}{9},
  \bibinfo{pages}{1735--1780}.
\bibitem[{Ji and Grishman(2011)}]{JiAndGrishman2011}
\bibinfo{author}{Ji, H.}, \bibinfo{author}{Grishman, R.}, \bibinfo{year}{2011}.
\newblock \bibinfo{title}{Knowledge base population: Successful approaches and
  challenges}, in: \bibinfo{booktitle}{Proceedings of the 49th Annual Meeting
  of the Association for Computational Linguistics}, pp.
  \bibinfo{pages}{1148--1158}.
\bibitem[{Johansen(2019)}]{johansen2019ner}
\bibinfo{author}{Johansen, B.}, \bibinfo{year}{2019}.
\newblock \bibinfo{title}{Named-entity recognition for norwegian}, in:
  \bibinfo{booktitle}{Proceedings of the 22nd Nordic Conference on
  Computational Linguistics, NoDaLiDa}.
\bibitem[{Jurafsky and Martin(2008)}]{SLP2008}
\bibinfo{author}{Jurafsky, D.}, \bibinfo{author}{Martin, J.},
  \bibinfo{year}{2008}.
\newblock \bibinfo{title}{Speech and Language Processing}.
\newblock \bibinfo{edition}{2nd} ed., \bibinfo{publisher}{Prentice Hall}.
\bibitem[{Jurafsky and Martin(2020)}]{SLP2020}
\bibinfo{author}{Jurafsky, D.}, \bibinfo{author}{Martin, J.},
  \bibinfo{year}{2020}.
\newblock \bibinfo{title}{Speech and Language Processing}.
\newblock \bibinfo{edition}{3nd ed. draft} ed.
\bibitem[{Krallinger et~al.(2015)Krallinger, Leitner, Rabal, Vazquez, Oyarzabal
  and Valencia}]{KrallingerEtal2013}
\bibinfo{author}{Krallinger, M.}, \bibinfo{author}{Leitner, F.},
  \bibinfo{author}{Rabal, O.}, \bibinfo{author}{Vazquez, M.},
  \bibinfo{author}{Oyarzabal, J.}, \bibinfo{author}{Valencia, A.},
  \bibinfo{year}{2015}.
\newblock \bibinfo{title}{Overview of the chemical compound and drug name
  recognition (chemdner) task}, in: \bibinfo{booktitle}{BioCreative Challenge
  Evaluation Workshop}, pp. \bibinfo{pages}{2--33}.
\bibitem[{Lafferty et~al.(2001)Lafferty, McCallum and Pereira}]{CRF2001}
\bibinfo{author}{Lafferty, J.}, \bibinfo{author}{McCallum, A.},
  \bibinfo{author}{Pereira, F.}, \bibinfo{year}{2001}.
\newblock \bibinfo{title}{Conditional random fields: Probabilistic models for
  segmenting and labeling sequence data}, in: \bibinfo{booktitle}{Proceedings
  of International Conference on Machine Learning}, pp.
  \bibinfo{pages}{281--289}.
\bibitem[{Li(1992)}]{Li1992}
\bibinfo{author}{Li, W.}, \bibinfo{year}{1992}.
\newblock \bibinfo{title}{Random texts exhibit zipf's-law-like word frequency}.
\newblock \bibinfo{journal}{IEEE Transactions on Information Theory}
  \bibinfo{volume}{38}, \bibinfo{pages}{1842--1845}.
\bibitem[{Li(2002)}]{Li2002}
\bibinfo{author}{Li, W.}, \bibinfo{year}{2002}.
\newblock \bibinfo{title}{Zipf's law everywhere}.
\newblock \bibinfo{journal}{Glottometrics} \bibinfo{volume}{5},
  \bibinfo{pages}{14--21}.
\bibitem[{Ling et~al.(2015)Ling, Singh and Weld}]{LingEtal2015}
\bibinfo{author}{Ling, X.}, \bibinfo{author}{Singh, S.}, \bibinfo{author}{Weld,
  D.S.}, \bibinfo{year}{2015}.
\newblock \bibinfo{title}{Design challenges for entity linking}.
\newblock \bibinfo{journal}{Transactions of the Association for Computational
  Linguistics} \bibinfo{volume}{3}, \bibinfo{pages}{315--328}.
\bibitem[{Ling and Weld(2012)}]{LingAndWeld2012}
\bibinfo{author}{Ling, X.}, \bibinfo{author}{Weld, D.S.}, \bibinfo{year}{2012}.
\newblock \bibinfo{title}{Fine-grained entity recognition}, in:
  \bibinfo{booktitle}{Proceedings of the Twenty-Sixth Conference on Artificial
  Intelligence}.
\bibitem[{Liu(2012)}]{Liu2012}
\bibinfo{author}{Liu, B.}, \bibinfo{year}{2012}.
\newblock \bibinfo{title}{Sentiment Analysis and Opinion Mining}.
\newblock \bibinfo{publisher}{Morgan \& Claypool Publishers}.
\bibitem[{Malone and Maher(2012)}]{MaloneMaher2012}
\bibinfo{author}{Malone, D.}, \bibinfo{author}{Maher, K.},
  \bibinfo{year}{2012}.
\newblock \bibinfo{title}{Investigating the distribution of password choices},
  in: \bibinfo{booktitle}{Proceedings of the 21th International Conference on
  World Wide Web}, pp. \bibinfo{pages}{301--310}.
\bibitem[{Manning and Schutze(1999)}]{FSNLP1999}
\bibinfo{author}{Manning, C.}, \bibinfo{author}{Schutze, H.},
  \bibinfo{year}{1999}.
\newblock \bibinfo{title}{Foundations of Statistical Natural Language
  Processing}.
\newblock \bibinfo{publisher}{Cambride: MIT Press}.
\bibitem[{Mazur and Dale(2010)}]{WikiWars2010}
\bibinfo{author}{Mazur, P.}, \bibinfo{author}{Dale, R.}, \bibinfo{year}{2010}.
\newblock \bibinfo{title}{Wikiwars: A new corpus for research on temporal
  expressions}, in: \bibinfo{booktitle}{Proceedings of the 2010 Conference on
  Empirical Methods in Natural Language Processing}, pp.
  \bibinfo{pages}{913--922}.
\bibitem[{Miller(1957)}]{Miller1957}
\bibinfo{author}{Miller, G.}, \bibinfo{year}{1957}.
\newblock \bibinfo{title}{Some effects of intermittent silence}.
\newblock \bibinfo{journal}{American Journal of Psychology}
  \bibinfo{volume}{70}, \bibinfo{pages}{311--314}.
\bibitem[{Miller(1965)}]{Miller1965}
\bibinfo{author}{Miller, G.}, \bibinfo{year}{1965}.
\newblock \bibinfo{title}{"Introduction" in The Psycho-biology of Language: An
  Introduction to Dynamic Philology (1935)}.
\newblock \bibinfo{publisher}{MIT Press}.
\bibitem[{Miller et~al.(1958)Miller, Newman and
  Friedman}]{MillerEtal1958Length}
\bibinfo{author}{Miller, G.A.}, \bibinfo{author}{Newman, E.B.},
  \bibinfo{author}{Friedman, E.A.}, \bibinfo{year}{1958}.
\newblock \bibinfo{title}{Length-frequency statistics for written english}.
\newblock \bibinfo{journal}{Information and Control} \bibinfo{volume}{1},
  \bibinfo{pages}{370--389}.
\bibitem[{Nadeau and Sekine(2007)}]{NERSurvey2007}
\bibinfo{author}{Nadeau, D.}, \bibinfo{author}{Sekine, S.},
  \bibinfo{year}{2007}.
\newblock \bibinfo{title}{A survey of named entity recognition and
  classification}.
\newblock \bibinfo{journal}{Lingvisticae Investigationes} \bibinfo{volume}{30},
  \bibinfo{pages}{3--26}.
\bibitem[{Nettasinghe and
  Krishnamurthy(2021)}]{NettasingheKrishnamurthy2021KDD}
\bibinfo{author}{Nettasinghe, B.}, \bibinfo{author}{Krishnamurthy, V.},
  \bibinfo{year}{2021}.
\newblock \bibinfo{title}{Maximum likelihood estimation of power-law degree
  distributions via friendship paradox-based sampling}.
\newblock \bibinfo{journal}{ACM Transactions on Knowledge Discovery from Data}
  \bibinfo{volume}{15}, \bibinfo{pages}{1--28}.
\bibitem[{Newman(2005)}]{PowerLaw2005}
\bibinfo{author}{Newman, M.E.}, \bibinfo{year}{2005}.
\newblock \bibinfo{title}{Power laws, pareto distributions and zipf's law}.
\newblock \bibinfo{journal}{Contemporary physics} \bibinfo{volume}{46},
  \bibinfo{pages}{323--351}.
\bibitem[{Paccosi and Aprosio(2021)}]{paccosi2021kind}
\bibinfo{author}{Paccosi, T.}, \bibinfo{author}{Aprosio, A.P.},
  \bibinfo{year}{2021}.
\newblock \bibinfo{title}{Kind: an italian multi-domain dataset for named
  entity recognition}.
\newblock \bibinfo{journal}{arXiv preprint arXiv:2112.15099} .
\bibitem[{P{\'e}rez-Casany and Casellas(2013)}]{perez2013marshall}
\bibinfo{author}{P{\'e}rez-Casany, M.}, \bibinfo{author}{Casellas, A.},
  \bibinfo{year}{2013}.
\newblock \bibinfo{title}{Marshall-olkin extended zipf distribution}.
\newblock \bibinfo{journal}{arXiv preprint arXiv:1304.4540} .
\bibitem[{Piantadosi(2014)}]{Piantadosi2014}
\bibinfo{author}{Piantadosi, S.T.}, \bibinfo{year}{2014}.
\newblock \bibinfo{title}{Zipf's word frequency law in natural language: A
  critical review and future directions}.
\newblock \bibinfo{journal}{Psychonomic bulletin \& review}
  \bibinfo{volume}{21}, \bibinfo{pages}{1112--1130}.
\bibitem[{Pontiki et~al.(2015)Pontiki, Galanis, Papageorgiou, Manandhar and
  Androutsopoulos}]{PontikiEtal2015}
\bibinfo{author}{Pontiki, M.}, \bibinfo{author}{Galanis, D.},
  \bibinfo{author}{Papageorgiou, H.}, \bibinfo{author}{Manandhar, S.},
  \bibinfo{author}{Androutsopoulos, I.}, \bibinfo{year}{2015}.
\newblock \bibinfo{title}{Semeval-2015 task 12: Aspect based sentiment
  analysis}, in: \bibinfo{booktitle}{Proceedings of the 9th International
  Workshop on Sementic Evaluation}, pp. \bibinfo{pages}{486--495}.
\bibitem[{Pontiki et~al.(2014)Pontiki, Galanis, Pavlopoulos, Papageorgiou,
  Androutsopoulos and Manandhar}]{PontikiEtal2014}
\bibinfo{author}{Pontiki, M.}, \bibinfo{author}{Galanis, D.},
  \bibinfo{author}{Pavlopoulos, J.}, \bibinfo{author}{Papageorgiou, H.},
  \bibinfo{author}{Androutsopoulos, I.}, \bibinfo{author}{Manandhar, S.},
  \bibinfo{year}{2014}.
\newblock \bibinfo{title}{Semeval-2014 task 4: Aspect based sentiment
  analysis}, in: \bibinfo{booktitle}{Proceedings of the 8th International
  Workshop on Semantic Evaluation}, pp. \bibinfo{pages}{27--35}.
\bibitem[{Pradhan et~al.(2013)Pradhan, Moschitti, Xue, Ng, Bjorkelund,
  Uryupina, Zhang and Zhong}]{OntoNotes2013}
\bibinfo{author}{Pradhan, S.}, \bibinfo{author}{Moschitti, A.},
  \bibinfo{author}{Xue, N.}, \bibinfo{author}{Ng, H.T.},
  \bibinfo{author}{Bjorkelund, A.}, \bibinfo{author}{Uryupina, O.},
  \bibinfo{author}{Zhang, Y.}, \bibinfo{author}{Zhong, Z.},
  \bibinfo{year}{2013}.
\newblock \bibinfo{title}{Towards robust linguistic analysis using ontonotes},
  in: \bibinfo{booktitle}{Proceedings of the 7th Conference on Computational
  Natural Language Learning}, pp. \bibinfo{pages}{143--152}.
\bibitem[{Pratap et~al.(2022)Pratap, Raja, Agarwal, Alzabut, Niezabitowski and
  Hincal}]{pratap2022further}
\bibinfo{author}{Pratap, A.}, \bibinfo{author}{Raja, R.},
  \bibinfo{author}{Agarwal, R.P.}, \bibinfo{author}{Alzabut, J.},
  \bibinfo{author}{Niezabitowski, M.}, \bibinfo{author}{Hincal, E.},
  \bibinfo{year}{2022}.
\newblock \bibinfo{title}{Further results on asymptotic and finite-time
  stability analysis of fractional-order time-delayed genetic regulatory
  networks}.
\newblock \bibinfo{journal}{Neurocomputing} \bibinfo{volume}{475},
  \bibinfo{pages}{26--37}.
\bibitem[{Pratap et~al.(2019)Pratap, Raja, Cao, Rajchakit and
  Fardoun}]{pratap2019stability}
\bibinfo{author}{Pratap, A.}, \bibinfo{author}{Raja, R.}, \bibinfo{author}{Cao,
  J.}, \bibinfo{author}{Rajchakit, G.}, \bibinfo{author}{Fardoun, H.M.},
  \bibinfo{year}{2019}.
\newblock \bibinfo{title}{Stability and synchronization criteria for fractional
  order competitive neural networks with time delays: An asymptotic expansion
  of mittag leffler function}.
\newblock \bibinfo{journal}{Journal of the Franklin Institute}
  \bibinfo{volume}{356}, \bibinfo{pages}{2212--2239}.
\bibitem[{Pustejovsky et~al.(2003a)Pustejovsky, Castano, Ingria, Sauri,
  Gaizauskas, Setzer, Katz and Radev}]{TimeML2003}
\bibinfo{author}{Pustejovsky, J.}, \bibinfo{author}{Castano, J.},
  \bibinfo{author}{Ingria, R.}, \bibinfo{author}{Sauri, R.},
  \bibinfo{author}{Gaizauskas, R.}, \bibinfo{author}{Setzer, A.},
  \bibinfo{author}{Katz, G.}, \bibinfo{author}{Radev, D.},
  \bibinfo{year}{2003}a.
\newblock \bibinfo{title}{Timeml: Robust specification of event and temporal
  expressions in text}.
\newblock \bibinfo{journal}{New Directions in Question Answering}
  \bibinfo{volume}{3}, \bibinfo{pages}{28--34}.
\bibitem[{Pustejovsky et~al.(2003b)Pustejovsky, Hanks, Sauri, See, Gaizauskas,
  Setzer, Sundheim, Radev, Day, Ferro and Lazo}]{TimeBank2003}
\bibinfo{author}{Pustejovsky, J.}, \bibinfo{author}{Hanks, P.},
  \bibinfo{author}{Sauri, R.}, \bibinfo{author}{See, A.},
  \bibinfo{author}{Gaizauskas, R.}, \bibinfo{author}{Setzer, A.},
  \bibinfo{author}{Sundheim, B.}, \bibinfo{author}{Radev, D.},
  \bibinfo{author}{Day, D.}, \bibinfo{author}{Ferro, L.},
  \bibinfo{author}{Lazo, M.}, \bibinfo{year}{2003}b.
\newblock \bibinfo{title}{The timebank corpus}.
\newblock \bibinfo{journal}{Corpus Linguistics} \bibinfo{volume}{2003},
  \bibinfo{pages}{647--656}.
\bibitem[{Ritter et~al.(2011)Ritter, Clark, Mausam and
  Etzioni}]{RitterEtal2011}
\bibinfo{author}{Ritter, A.}, \bibinfo{author}{Clark, S.},
  \bibinfo{author}{Mausam}, \bibinfo{author}{Etzioni, O.},
  \bibinfo{year}{2011}.
\newblock \bibinfo{title}{Named entity recognition in tweets: An experimental
  study}, in: \bibinfo{booktitle}{Proceedings of the 2011 Conference on
  Empirical Methods in Natural Language Processing}, pp.
  \bibinfo{pages}{1524--1534}.
\bibitem[{Sang and Meulder(2003)}]{CoNLL2003}
\bibinfo{author}{Sang, E.F.T.K.}, \bibinfo{author}{Meulder, F.D.},
  \bibinfo{year}{2003}.
\newblock \bibinfo{title}{Introduction to the conll-2003 shared task:
  Language-independent named entity recognition}, in:
  \bibinfo{booktitle}{Proceedings of the 7th Conference on Natural Language
  Learning}, pp. \bibinfo{pages}{142--147}.
\bibitem[{Sigurd et~al.(2004)Sigurd, Eeg-Olofsson and van~de
  Weijer}]{GigurdEtal2004}
\bibinfo{author}{Sigurd, B.}, \bibinfo{author}{Eeg-Olofsson, M.},
  \bibinfo{author}{van~de Weijer, J.}, \bibinfo{year}{2004}.
\newblock \bibinfo{title}{Word length, sentence length and frequency - zipf
  revisited}.
\newblock \bibinfo{journal}{Studia Linguistica} \bibinfo{volume}{58},
  \bibinfo{pages}{37--52}.
\bibitem[{Smirnov(1948)}]{smirnov1948table}
\bibinfo{author}{Smirnov, N.}, \bibinfo{year}{1948}.
\newblock \bibinfo{title}{Table for estimating the goodness of fit of empirical
  distributions}.
\newblock \bibinfo{journal}{The annals of mathematical statistics}
  \bibinfo{volume}{19}, \bibinfo{pages}{279--281}.
\bibitem[{Stephens(1974)}]{stephens1974edf}
\bibinfo{author}{Stephens, M.A.}, \bibinfo{year}{1974}.
\newblock \bibinfo{title}{Edf statistics for goodness of fit and some
  comparisons}.
\newblock \bibinfo{journal}{Journal of the American statistical Association}
  \bibinfo{volume}{69}, \bibinfo{pages}{730--737}.
\bibitem[{Strauss et~al.(2016)Strauss, Toma, Ritter, de~Marneffe and
  Xu}]{WNUT16}
\bibinfo{author}{Strauss, B.}, \bibinfo{author}{Toma, B.E.},
  \bibinfo{author}{Ritter, A.}, \bibinfo{author}{de~Marneffe, M.C.},
  \bibinfo{author}{Xu, W.}, \bibinfo{year}{2016}.
\newblock \bibinfo{title}{Results of the wnut16 named entity recognition shared
  task}, in: \bibinfo{booktitle}{Proceedings of the 2nd Workshop on Noisy
  User-generated Text}, pp. \bibinfo{pages}{138--144}.
\bibitem[{Takeuchi and Collier(2005)}]{TakeuchiAndCollier2005}
\bibinfo{author}{Takeuchi, K.}, \bibinfo{author}{Collier, N.},
  \bibinfo{year}{2005}.
\newblock \bibinfo{title}{Bio-medical entity extraction using support vector
  machines}.
\newblock \bibinfo{journal}{Artificial Intelligence In Medicine}
  \bibinfo{volume}{33}, \bibinfo{pages}{125--137}.
\bibitem[{UzZaman et~al.(2013)UzZaman, Llorens, Derczynski, Verhagen, Allen and
  Pustejovsky}]{TempEval-3}
\bibinfo{author}{UzZaman, N.}, \bibinfo{author}{Llorens, H.},
  \bibinfo{author}{Derczynski, L.}, \bibinfo{author}{Verhagen, M.},
  \bibinfo{author}{Allen, J.}, \bibinfo{author}{Pustejovsky, J.},
  \bibinfo{year}{2013}.
\newblock \bibinfo{title}{Semeval-2013 task 1: Tempeval-3: Evaluating time
  expressions, events, and temporal relations}, in:
  \bibinfo{booktitle}{Proceedings of the 7th International Workshop on Semantic
  Evaluation}, pp. \bibinfo{pages}{1--9}.
\bibitem[{Wake(1957)}]{Wake1957}
\bibinfo{author}{Wake, W.C.}, \bibinfo{year}{1957}.
\newblock \bibinfo{title}{Sentence-length distributions of greek authors}.
\newblock \bibinfo{journal}{Journal of the Royal Statistical Society: Series A
  (General)} \bibinfo{volume}{120}, \bibinfo{pages}{331--346}.
\bibitem[{Wang et~al.(2017)Wang, Cheng, Wang, Huang and
  Jian}]{WangEtal2017Zipf}
\bibinfo{author}{Wang, D.}, \bibinfo{author}{Cheng, H.}, \bibinfo{author}{Wang,
  P.}, \bibinfo{author}{Huang, X.}, \bibinfo{author}{Jian, G.},
  \bibinfo{year}{2017}.
\newblock \bibinfo{title}{Zipf's law in passwords}.
\newblock \bibinfo{journal}{IEEE Transactions on Information Forensics and
  Security} \bibinfo{volume}{12}, \bibinfo{pages}{2776--2791}.
\bibitem[{Wang et~al.(2020)Wang, Song, Li, Guan and Han}]{COVID19NER}
\bibinfo{author}{Wang, X.}, \bibinfo{author}{Song, X.}, \bibinfo{author}{Li,
  B.}, \bibinfo{author}{Guan, Y.}, \bibinfo{author}{Han, J.},
  \bibinfo{year}{2020}.
\newblock \bibinfo{title}{Comprehensive named entity recognition on cord-19
  with distant or weak supervision}, in: \bibinfo{booktitle}{arXiv preprint:
  arxiv.org/abs/2003.12218}.
\bibitem[{Weischedel and Brunstein(2005)}]{BBN2005}
\bibinfo{author}{Weischedel, R.}, \bibinfo{author}{Brunstein, A.},
  \bibinfo{year}{2005}.
\newblock \bibinfo{title}{Bbn pronoun coreference and entity type corpus}.
\newblock \bibinfo{journal}{Linguistic Data Consortium} \bibinfo{volume}{112}.
\bibitem[{Williams(1940)}]{Williams1940}
\bibinfo{author}{Williams, C.B.}, \bibinfo{year}{1940}.
\newblock \bibinfo{title}{A note on the statistical analysis of sentence-length
  as a criterion of literary style}.
\newblock \bibinfo{journal}{Biometrika} \bibinfo{volume}{31},
  \bibinfo{pages}{356--361}.
\bibitem[{Williams(1975)}]{Williams1975Mendenhall}
\bibinfo{author}{Williams, C.B.}, \bibinfo{year}{1975}.
\newblock \bibinfo{title}{Mendenhall's studies of word-length distribution in
  the works of shakespeare and bacon}.
\newblock \bibinfo{journal}{Biometrika} \bibinfo{volume}{62},
  \bibinfo{pages}{207--212}.
\bibitem[{Wimmer et~al.(1994)Wimmer, Kohler, Grotjahn and
  Altmann}]{WimmerEtal1994}
\bibinfo{author}{Wimmer, G.}, \bibinfo{author}{Kohler, R.},
  \bibinfo{author}{Grotjahn, R.}, \bibinfo{author}{Altmann, G.},
  \bibinfo{year}{1994}.
\newblock \bibinfo{title}{Towards a theory of word length distribution}.
\newblock \bibinfo{journal}{Journal of Quantitative Linguistics}
  \bibinfo{volume}{1}, \bibinfo{pages}{98--106}.
\bibitem[{Yadav and Bethard(2018)}]{NER2018Survey}
\bibinfo{author}{Yadav, V.}, \bibinfo{author}{Bethard, S.},
  \bibinfo{year}{2018}.
\newblock \bibinfo{title}{A survey on recent advances in named entity
  recognition from deep learning models}, in: \bibinfo{booktitle}{Proceedings
  of the 27th International Conference on Computational Linguistics},
  \bibinfo{publisher}{Association for Computational Linguistics},
  \bibinfo{address}{Santa Fe, New Mexico, USA}. pp.
  \bibinfo{pages}{2145--2158}.
\bibitem[{Zhong(2020)}]{MyThesis2020}
\bibinfo{author}{Zhong, X.}, \bibinfo{year}{2020}.
\newblock \bibinfo{title}{Time Expression and Named Entity Analysis and
  Recognition}.
\newblock Ph.D. thesis. Nanyang Technological University, Singapore.
\bibitem[{Zhong and Cambria(2018)}]{TOMN2018}
\bibinfo{author}{Zhong, X.}, \bibinfo{author}{Cambria, E.},
  \bibinfo{year}{2018}.
\newblock \bibinfo{title}{Time expression recognition using a constituent-based
  tagging scheme}, in: \bibinfo{booktitle}{Proceedings of the 2018 World Wide
  Web Conference}, pp. \bibinfo{pages}{983--992}.
\bibitem[{Zhong and Cambria(2023)}]{Zhong2023Time}
\bibinfo{author}{Zhong, X.}, \bibinfo{author}{Cambria, E.},
  \bibinfo{year}{2023}.
\newblock \bibinfo{title}{Time expression recognition and normalization: A
  survey}.
\newblock \bibinfo{journal}{Artificial Intelligence Review}
  \bibinfo{volume}{56}, \bibinfo{pages}{9115--9140}.
\bibitem[{Zhong et~al.(2020)Zhong, Cambria and Hussain}]{UGTO2020}
\bibinfo{author}{Zhong, X.}, \bibinfo{author}{Cambria, E.},
  \bibinfo{author}{Hussain, A.}, \bibinfo{year}{2020}.
\newblock \bibinfo{title}{Extracting time expressions and named entities with
  constituent-based tagging schemes}.
\newblock \bibinfo{journal}{Cognitive Computation} \bibinfo{volume}{12},
  \bibinfo{pages}{844--862}.
\bibitem[{Zhong et~al.(2022a)Zhong, Cambria and Hussain}]{ZhongEtal2022syntax}
\bibinfo{author}{Zhong, X.}, \bibinfo{author}{Cambria, E.},
  \bibinfo{author}{Hussain, A.}, \bibinfo{year}{2022}a.
\newblock \bibinfo{title}{Does semantics aid syntax? an empirical study on
  named entity recognition and classification}.
\newblock \bibinfo{journal}{Neural Computing and Applications}
  \bibinfo{volume}{34}, \bibinfo{pages}{8373--8384}.
\bibitem[{Zhong et~al.(2017)Zhong, Sun and Cambria}]{SynTime2017}
\bibinfo{author}{Zhong, X.}, \bibinfo{author}{Sun, A.},
  \bibinfo{author}{Cambria, E.}, \bibinfo{year}{2017}.
\newblock \bibinfo{title}{Time expression analysis and recognition using
  syntactic token types and general heuristic rules}, in:
  \bibinfo{booktitle}{Proceedings of the 55th Annual Meeting of the Association
  for Computational Linguistics}, pp. \bibinfo{pages}{420--429}.
\bibitem[{Zhong et~al.(2022b)Zhong, Wang and Zhang}]{zhong2022least}
\bibinfo{author}{Zhong, X.}, \bibinfo{author}{Wang, M.},
  \bibinfo{author}{Zhang, H.}, \bibinfo{year}{2022}b.
\newblock \bibinfo{title}{Is least-squares inaccurate in fitting power-law
  distributions? the criticism is complete nonsense}, in:
  \bibinfo{booktitle}{Proceedings of the ACM Web Conference 2022}, pp.
  \bibinfo{pages}{2748--2758}.
\bibitem[{Zipf(1936)}]{Zipf1936}
\bibinfo{author}{Zipf, G.}, \bibinfo{year}{1936}.
\newblock \bibinfo{title}{The Psychobiology of Language}.
\newblock \bibinfo{publisher}{London: Routledge}.
\bibitem[{Zipf(1949)}]{Zipf1949}
\bibinfo{author}{Zipf, G.}, \bibinfo{year}{1949}.
\newblock \bibinfo{title}{Human Behavior and the Principle of Least Effort: An
  Introduction to Human Ecology}.
\newblock \bibinfo{publisher}{Addison-Wesley Press, Inc.}

\end{thebibliography}

\end{document}